\documentclass{article}
\usepackage{arxiv}
\usepackage[T1]{fontenc}
\usepackage[utf8]{inputenc}
\usepackage{authblk}

\usepackage{lineno,hyperref}

\usepackage{amssymb}
\usepackage{amsmath}
\usepackage{amsfonts}

\usepackage{algorithmic}
\usepackage{algorithm}

\usepackage{eqparbox}

\usepackage{etoolbox}

\usepackage{color}
\usepackage{multirow}
\usepackage{booktabs}

\usepackage{graphicx}
\usepackage{subfig}

\usepackage{appendix}

\usepackage[numbers]{natbib}

\modulolinenumbers[5]

\newcommand\fk[1]{\textcolor{black}{#1}}

\title{End-to-end Kernel Learning\\ via Generative Random Fourier Features}

\author[1]{Kun Fang}
\author[2]{Fanghui Liu}
\author[1]{Xiaolin Huang}
\author[1]{Jie Yang}
\affil[1]{Department of Automation, Shanghai Jiao Tong University\\
{\tt\small
\{fanghenshao,xiaolinhuang,jieyang\}@sjtu.edu.cn}}

\affil[2]{LIONS Lab, \'{E}cole Polytechnique F\'{e}d\'{e}rale de Lausanne (EPFL), Switzerland\\
{\tt\small fanghui.liu@epfl.ch}}

\date{}

\begin{document}

\maketitle

\begin{abstract}
Random Fourier features (RFFs) provide a promising way for kernel learning in a spectral case.
Current RFFs-based kernel learning methods usually work in a two-stage way. 
In the first-stage process, learning an optimal feature map is often formulated as a target alignment problem, which aims to align the learned kernel with a pre-defined target kernel (usually the ideal kernel). 
In the second-stage process, a linear learner is conducted with respect to the mapped random features.
Nevertheless, the pre-defined kernel in target alignment is not necessarily optimal for the generalization of the linear learner. 
Instead, in this paper, we consider a one-stage process that incorporates the kernel learning and linear learner into a unifying framework. 
To be specific, a generative network via RFFs is devised to implicitly learn the kernel, followed by a linear classifier parameterized as a full-connected layer.
Then the generative network and the classifier are jointly trained by solving an empirical risk minimization (ERM) problem to reach a one-stage solution.
This end-to-end scheme naturally allows deeper features, in correspondence to a multi-layer structure, and shows superior generalization performance over the classical two-stage, RFFs-based methods in real-world classification tasks.
Moreover, inspired by the randomized resampling mechanism of the proposed method, its enhanced adversarial robustness is investigated and experimentally verified.

\end{abstract}

\section{Introduction}
\label{sec-introduction}
Kernel methods reveal the non-linear property hidden in data and have been extensively studied in recent decades \cite{scholkopf2001learning,filippone2008survey}. 
The selection of the kernel still remains a non-trivial problem, which requires prior knowledge and directly affects the algorithm performance. 
Hence, learning a suitable kernel from data has been a universal choice \cite{nazarpour2015two,lauriola2020enhancing}.
In recent decades, a series of researches have been devoted to exploit the random Fourier features (RFFs) \cite{rahimi2008random} in the kernel learning task \cite{sinha2016learning,li2019implicit}.

The pioneering work \cite{rahimi2008random} of RFFs constructed an explicit feature map $\phi$ to approximate a positive definite, shift-invariant kernel.
The feature map $\phi$ maps the input point $\mathbf{x}$ to the so-called random Fourier features $\phi(\mathbf{x},\mathbf{w})$, where the associated weights $\mathbf{w}$ are sampled from the spectral distribution of the kernel function.
A linear learner is then conducted on the mapped features to categorize them.
Intuitively, RFFs allow us to learn the weights $\mathbf{w}$ in a spectral sense, i.e., to learn a kernel  \cite{sinha2016learning,li2019implicit}. 
To be specific, 
to involve the data information, typical approaches \cite{sinha2016learning,li2019implicit} proposed to solve the following target alignment problem \cite{cristianini2002kernel} to learn the weights $\mathbf{w}$,
\begin{equation}
\mathop{\arg \max}_{\mathbf{w}} \ \sum_{i,j}
y_iy_j \phi(\mathbf{x}_i,\mathbf{w}) ^T \phi(\mathbf{x}_j,\mathbf{w}),
\label{op-KA}
\end{equation}
where $(\mathbf{x}_i,y_i)$ and $(\mathbf{x}_j,y_j)$ denote the $i$-th and the $j$-th training samples respectively.
The inner product of the two mapped features $\phi(\mathbf{x}_i,\mathbf{w}) ^T\phi(\mathbf{x}_j,\mathbf{w})$ denotes the implicit kernel function value $k(\mathbf{x}_i,\mathbf{x}_j)$.
\citet{sinha2016learning} learned an optimal weight subset of $\mathbf{w}$ by solving Eq.\eqref{op-KA} for an optimal feature subset of the vanilla RFFs, and proved the consistency and generalization bound.
The proposed algorithm \cite{sinha2016learning} is efficient and highly scalable.
\citet{li2019implicit} incorporated a deep neural network to implicitly learn the kernel, and trained the network by solving Eq.\eqref{op-KA}.
The work in \cite{li2019implicit} shows that the performance could be improved by involving a network to model the kernel distribution.

These kernel learning methods \cite{sinha2016learning,li2019implicit}  follow a \emph{two-stage} scheme: They first learn the random features $\phi(\mathbf{x},\mathbf{w})$, which is equivalent to learning the weights $\mathbf{w}$ since the map $\phi$ is explicit, by solving Eq.\eqref{op-KA}, and then learn a linear classifier on the features.
The optimization target of Eq.\eqref{op-KA} is to align the learned kernel with the pre-defined ideal kernel $\mathbf{y}\mathbf{y}^T$.
However, this commonly-used ideal kernel in target alignment is not necessarily optimal for the generalization of the linear classifier learned in the second stage.
Hence, in such a two-stage way, the random features learned in the first stage perhaps show excellent approximation performance towards the ideal kernel, but do not take much care of the generalization or classification performance, which could be further improved.

To address this issue, towards learning a more well-generalized kernel, we propose to jointly learn the random features and the classifier in an end-to-end way by straightly solving the following expectation risk minimization problem,
\begin{equation}
\mathop{\arg \min}_{G,C} \  \mathbb{E}_{(\mathbf{x},y)}L\Big(C\big(\phi\big(\mathbf{x},\mathbb{E}_{\mathbf{n}}G(\mathbf{n})\big)\big), y\Big),
\label{op-ExpRM}
\end{equation}
where $L(\cdot,\cdot)$ denotes the loss function and $\mathbf{n}$ denotes the noise random variable.  
A generative network $G(\cdot)$, a.k.a. a generator, is devised to learn the kernel distribution, which takes a set of noises as input and samples weights from the learned kernel distribution.
Then, the corresponding random features are constructed based on the generated weights according to the feature map $\phi$.
A linear classifier $C(\cdot)$ parameterized as a full-connected (FC) layer is applied to the random features to categorize them.
The generative network and linear classifier are jointly trained by directly minimizing the loss between the true labels and predicted ones.
Therefore, we achieve an end-to-end, \emph{one-stage} solution, which no longer pursues the approximation ability of random features towards any pre-defined kernels.
Instead, it is expected that distributions of the underlying kernel can be modeled by the generative network for better classification or generalization performance.
The proposed method is called Generative RFFs (GRFFs) since the random features are built via the generative network.

Besides, this end-to-end training strategy naturally allows deeper features, in correspondence to a multi-layer structure of GRFFs.
An associated progressively training strategy is designed to empirically guarantee the convergence of the multi-layer GRFFs.
The multi-layer structure further boosts the generalization performance, since a good kernel on features is learned in this way.
Moreover, the randomized resampling mechanism of GRFFs yields the non-deterministic weights and brings the merit of adversarial robustness gains, which is investigated and experimentally verified.

Our contributions can be summarized as follows,
\begin{itemize}
	\item To the best of our knowledge, this is the first work in RFF-based kernel learning algorithms to incorporate the kernel learning and the linear learner into a unifying framework via generative models for an end-to-end and one-stage solution. Empirical results indicate the superior generalization performance over the classical two-stage, RFFs-based methods.
	\item The end-to-end strategy enables us to employ a multi-layer structure for GRFFs, indicating a good kernel on deeper features. A progressively training strategy is devised to efficiently train the multiple generators. The multi-layer structure further advances the generalization performance.
	\item  The adversarial robustness of the proposed method is also investigated. Empirical results show that, to some degree, the randomized resampling mechanism associated with the learned distributions can alleviate the performance decrease against adversarial attacks.
\end{itemize}
The rest of this paper is organized as follows.
We briefly introduce the random Fourier features and the generative models in Section \ref{sec-preliminaries}.
The one-stage framework, multi-layer structure, and  progressively training strategy of the proposed method are explained in detail in Section \ref{sec-proposedmethods}.
Experiment results on classification tasks and adversarial robustness are shown in Section \ref{sec-experimentsresults}.
The discussions and the conclusions are drawn in Section \ref{sec-conclusion}.

\section{Preliminaries}
\label{sec-preliminaries}

\subsection{Random Fourier Features}
In kernel methods, a positive definite kernel $k(\mathbf{x}, \mathbf{x^\prime})$ with $\mathbf{x},\mathbf{x^\prime} \in \mathcal{R}^d$ defines a map $\Phi:\mathcal{R}^d \to \mathcal{H}$, which satisfies $k(\mathbf{x}, \mathbf{x^\prime})=\langle\Phi(\mathbf{x}), \Phi(\mathbf{x^\prime})\rangle_\mathcal{H}$, where $\langle \cdot,\cdot \rangle_\mathcal{H}$ denotes the inner product in a Reproducing Kernel Hilbert Space $\mathcal{H}$.
However, in large-scale kernel machines, there exist unacceptable high computation and memory costs: $\mathcal{O}(n^2)$ kernel evaluations, $\mathcal{O}(n^2)$ in space, and even $\mathcal{O}(n^3)$ in time to compute the inverse of the kernel matrix.
Therefore, randomized features are introduced in large-scale cases to approximate the kernel function \cite{rahimi2008random}.
The theoretical foundation is based on the Bochner's theorem \cite{rudin1962fourier}, which indicates that the Fourier transform of a kernel function is associated to a probability distribution:
A continuous and shift-invariant kernel $k(\mathbf{x},\mathbf{x'})=k(\mathbf{x}-\mathbf{x'})$ on $\mathcal{R}^d$ is positive definite if and only if $k(\cdot)$ is the Fourier transform of a non-negative measure $p(\mathbf{w})$. That is,
\begin{equation}
k(\mathbf{x}-\mathbf{x'})=\int_{\mathcal{R}^d}p(\mathbf{w})e^{\mathrm{j}\mathbf{w}^T(\mathbf{x}-\mathbf{x^\prime})}\mathrm{d}\mathbf{w}.
\label{eq-bochner}
\end{equation}
Considering a real-valued, shift-invariant and positive definite kernel $k$, by applying the Euler's formula $e^{\mathrm{j}x}=\cos(x)+\mathrm{j}\sin(x)$ to Eq.\eqref{eq-bochner}, we have
\begin{equation}
\begin{split}
k(\mathbf{x}-\mathbf{x'}) 
&=\mathrm{Re}[k(\mathbf{x}-\mathbf{x'})] =\int_{\mathcal{R}^d}p(\mathbf{w})\cos(\mathbf{w}^T(\mathbf{x}-\mathbf{x'}))\mathrm{d}\mathbf{w}.\\
= & \mathop{\int}\limits_{\mathcal{R}^d}p(\mathbf{w})
(\cos(\mathbf{w}^T\mathbf{x})\cos(\mathbf{w}^T\mathbf{x'})
+\sin(\mathbf{w}^T\mathbf{x})\sin(\mathbf{w}^T\mathbf{x'}))\mathrm{d}\mathbf{w}\\
= & \ \mathbb{E}_{\mathbf{w}}
\Big[
\big[\cos(\mathbf{w}^T\mathbf{x})\ \sin(\mathbf{w}^T\mathbf{x})\big]^T
\big[\cos(\mathbf{w}^T\mathbf{x'})\ \sin(\mathbf{w}^T\mathbf{x'})\big]
\Big].
\end{split}
\end{equation}
Therefore, one can construct an explicit feature map $\phi: \mathcal{R}^d \to \mathcal{R}^{2D}$ as follows:
\begin{equation}
\phi(\mathbf{x},\{\mathbf{w}_i\}^D_{i=1})
\equiv 
\sqrt{\frac{1}{D}}\big[\cos(\mathbf{w}^T_1\mathbf{x}) \cdots \cos(\mathbf{w}^T_D\mathbf{x})\  \sin(\mathbf{w}^T_1\mathbf{x}) \cdots \sin(\mathbf{w}^T_D\mathbf{x})\big],
\label{eq-RFF}
\end{equation}
known as the random Fourier features \cite{rahimi2008random}.
The set of weights $\{\mathbf{w}_i\}^D_{i=1}$ is sampled from the spectral distribution of the kernel function.
% Hence it could be observed that tuning the weight parameters $\{\mathbf{w}_i\}^D_{i=1}$ is equal to learning the kernel distribution.
The mapped random Fourier features satisfy 
$\phi(\mathbf{x},\{\mathbf{w}_i\}^D_{i=1})^T\phi(\mathbf{x^\prime},\{\mathbf{w}_i\}^D_{i=1}) 
\approx k(\mathbf{x}, \mathbf{x^\prime})$ \cite{rahimi2008random}. 
A detailed analysis of the convergence can be found in \cite{mohri2018foundations}.

The vanilla RFFs \cite{rahimi2008random} accelerate the large-scale kernel machines a lot and are data-independent.
Afterwards, RFFs have been widely utilized, including the kernel approximation \cite{liu2019random}, extreme learning machines \cite{zhang2019extreme}, the bias-variance trade-off in machine learning \cite{belkin2019reconciling} and the kernel learning task \cite{sinha2016learning,li2019implicit}.
A systematic survey on RFFs-based algorithms can be found in \cite{surveyRFF}.

Inspired by the vanilla RFFs, learning the features is equal to learning the weights $\{\mathbf{w}_i\}^D_{i=1}$, while learning the weights is equal to learning the kernel distribution.
Hence, to efficiently learn the distribution, generative models are introduced.

\subsection{Generative Models}
Generative models have been widely applied in learning distributions from data \cite{goodfellow2016deep}.
Various generative models can be categorized into two types: models that perform explicit probability density estimations, and models that perform as a \emph{sampler} sampling from the estimated distribution without a precise probability function.

Suppose a training set includes samples from a distribution $\mathbb{P}_{data}$, then the first type of generative models returns an estimation $\mathbb{P}_{model}$ of $\mathbb{P}_{data}$.
Given a particular value as input, the estimation $\mathbb{P}_{model}$ will output the corresponding probability.
While the second type tries to learn the latent $\mathbb{P}_{data}$ as well, but in a rather different way: The learned model actually simulates a sampling process, through which one can generate more samples from the estimated distribution.

There exist lots of researches in the second type of generative models using a deep neural network as the sampler.
In computer vision, the family of generative adversarial networks (GANs) \cite{goodfellow2014generative} has shown great generality in various situations \cite{radford2015unsupervised,karras2018progressive}.
The generative network in GANs performs as a useful image generator, which generates images by sampling from the learned image distribution.
Besides, in Bayesian deep learning, there is a series of hypernetwork-based works, which use a neural network, called hypernet, to generate the parameters in another neural network, called primary net.
The hypernet also performs as a sampler to learn the parameter posterior distribution of the primary net \cite{ukai2018hypernetwork,ratzlaff2019hypergan}.

Therefore, the generative networks are employed in the proposed GRFFs due to the powerful distribution learning ability. The resulting merits are two-fold: On the one hand, the GRFFs could learn a well-generalized kernel distribution from data.
On the other hand, the generative networks could produce non-deterministic weights, bringing the adversarial robustness gains.

\section{Generative Random Fourier Features Model}
\label{sec-proposedmethods}

\subsection{One-stage Generative Random Fourier Features}

Previous RFFs-based kernel learning methods construct random features via sampling from the spectral distribution of the kernel \cite{rahimi2008random} or via approximating the ideal kernel \cite{sinha2016learning,li2019implicit}, 
and then train a classifier on these features.
The classification performance of the random features learned in this two-stage manner can perhaps be further improved.
Therefore, we propose generative RFFs to jointly learn the features and the classifier by optimizing the expectation risk minimization problem in an end-to-end manner, which leads to a one-stage solution with better classification performance.

\begin{figure*}[!t]
	\centering
	\includegraphics[scale=0.9]{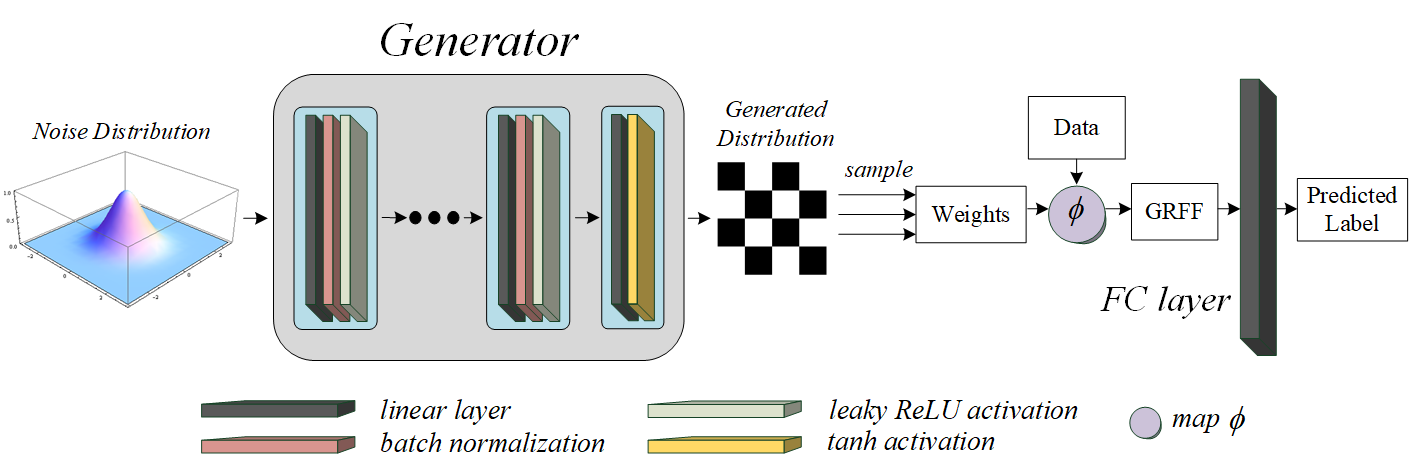}
	\caption{An illustration of the one-stage generative random Fourier features.}
	\label{fig_slgrff}
\end{figure*}

We first describe a general framework of our one-stage model, illustrated in Fig.\ref{fig_slgrff}.
A generator is designed to learn and to sample from the latent kernel distribution.
Then, the generative random Fourier features of the original data are constructed by the sampled weights from the learned distribution.
Finally, a linear classifier is applied to the features to categorize them.

The generator in our method performs as a \emph{sampler}.
It implicitly learns some distribution $\mathbb{P}_k$ and generates samples from it.
Given an arbitrary noise distribution $\mathbb{P}_0$ 
and a set of noises $\mathbf{N} = \{\mathbf{n}_i\}^D_{i=1}$ sampled from $\mathbb{P}_0$, 
the generator $\Phi_G$ takes them as input and generates the corresponding set of weights $\{\mathbf{w}_i\}^D_{i=1}$ sampled from $\mathbb{P}_k$:
\begin{equation}
\mathbf{w}_i = \Phi_G(\mathbf{n}_i), \mathbf{n}_i\sim \mathbb{P}_0, \mathbf{w}_i\sim \mathbb{P}_k, i=1,...,D.
\label{eq-generator}
\end{equation}

Given a training set $\{(\mathbf{x}_i, y_i)\}_{i=1}^n$,
the generative random Fourier features $\mathbf{z}_i$ of each training point $\mathbf{x}_i$ will be constructed with full use of the $D$ weights $\{\mathbf{w}_j\}^D_{j=1}$ according to Eq.\eqref{eq-RFF}:
\begin{equation}
\mathbf{z}_i=\phi(\mathbf{x}_i, \{\mathbf{w}_j\}^D_{j=1}), \mathbf{w}_j\sim \mathbb{P}_k, i=1,...,n.
\label{eq-GRFF}
\end{equation}
Notice that the weight $\mathbf{w}_j$ and the data point $\mathbf{x}_i$ are of the same dimension and that the dimension of $\mathbf{z}_i$ equals to $2D$.
In the end, the linear classifier $\Phi_C$ will be applied on the GRFFs $\mathbf{z}_i$ to predict the label $\hat{y}_i=\Phi_C(\mathbf{z}_i)$.

Denote $\theta_G$ and $\theta_C$ as the parameters of the generator $\Phi_G$ and the linear classifier $\Phi_C$ respectively. 
Combining Eq.\eqref{eq-generator} and Eq.\eqref{eq-GRFF},
we have the following empirical risk minimization (ERM) problem:
\begin{equation}
\mathop{\min}_{\theta_G,\theta_C}
\frac{1}{n}
\sum_{i=1}^{n}
L\Big(
\Phi_C\big(\phi\big(\mathbf{x}_i, \Phi_G(\mathbf{N})\big)\big)
, y_i\Big).
\label{loss-EmpRM}
\end{equation}

\subsection{Multi-layer Generative Random Fourier Features}

\begin{figure}[!t]
	\centering
	\includegraphics[scale=0.9]{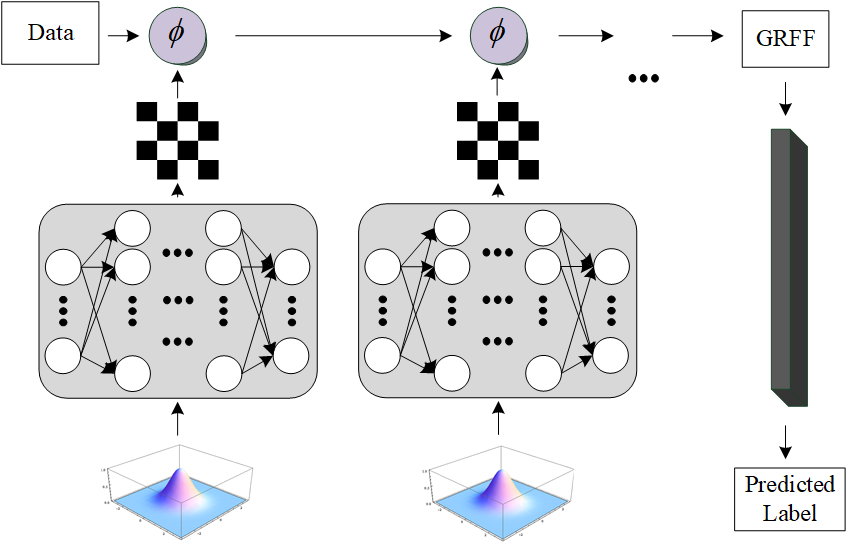}
	\caption{An illustration of the multi-layer structure of GRFFs.}
	\label{fig_mlgrff}
\end{figure}

The end-to-end training strategy by straightly optimizing the ERM problem naturally allows us to employ a multi-layer structure of GRFFs, shown in Fig.\ref{fig_mlgrff}.
To be specific, with one generator, we can build the random features of the original data, which can be viewed as the first layer of features.
Then, with another generator, a second layer of random features can be constructed in the same way based on the features from the previous layer.
After such a layer-by-layer operation, the random features in the last layer are followed with an FC layer.
The total networks, i.e., all the generators and the last FC layer, are again jointly trained to solve the ERM problem.
In this way, in each layer, there exists a corresponding generator which models the specific distribution on this layer of features, indicating a good kernel learned on features.

In those two-stage approaches \cite{rahimi2008random,sinha2016learning,li2019implicit}, the distribution is associated with a pre-defined kernel, which is restricted to a \emph{single-layer} structure, since the kernel for the multi-layer case is not clear.
By contrast, we cover the \emph{multi-layer} case, and thus have an advantage in learning a much more linear-separable pattern from data under the guidance of the learned kernels on features.

\fk{Without loss of generality, a detailed elaboration on the two-layer structure is presented below as an example, which could be naturally generalized to any number of layers (see Alg.\ref{alg-train}).} 
% Take the simplest two-layer structure as an example.
Denote $\Phi_{G_1}$ and $\Phi_{G_2}$ as the two generators in the first and second layers respectively.
The two generators, $\Phi_{G_1}$ and $\Phi_{G_2}$, take two sets of noises 
$\mathbf{N}_1=\{\mathbf{n}_i^1\}_{i=1}^{D_1}$ and $\mathbf{N}_2=\{\mathbf{n}_i^2\}_{i=1}^{D_2}$ independently sampled from the same distribution $\mathbb{P}_0$ as input respectively,
and generate the corresponding weights $\mathbf{W}_1=\{\mathbf{w}_i^1\}_{i=1}^{D_1}$ and  $\mathbf{W}_2=\{\mathbf{w}_i^2\}_{i=1}^{D_2}$:
\begin{equation}
\begin{split}
\mathbf{w}_i^1 = \Phi_{G_1}(\mathbf{n}_i^1), \mathbf{n}_i^1\sim \mathbb{P}_0, 
\mathbf{w}_i^1\sim \mathbb{P}_{k_1},
i=1,...,D_1,\\
\mathbf{w}_i^2 = \Phi_{G_2}(\mathbf{n}_i^2), \mathbf{n}_i^2\sim \mathbb{P}_0, 
\mathbf{w}_i^2\sim \mathbb{P}_{k_2},
i=1,...,D_2,
\end{split}
\label{eq-generator-ml}
\end{equation}
where $\mathbb{P}_{k_1}$ and $\mathbb{P}_{k_2}$ are the distributions modeled by $\Phi_{G_1}$ and $\Phi_{G_2}$ respectively.

In the first layer, a training sample $\mathbf{x}_i$ cooperates together with $\mathbf{W}_1$ to construct the generative random Fourier feature $\mathbf{z}_i^1$ 
according to Eq.\eqref{eq-RFF}:
\begin{equation}
\mathbf{z}_i^1=
\phi(\mathbf{x}_i, \{\mathbf{w}_j^1\}^{D_1}_{j=1}), 
\mathbf{w}_j^1\sim \mathbb{P}_{k_1},
i=1,...,n.
\label{eq-GRFF-layer1}
\end{equation}
Then, the random feature $\mathbf{z}_i^1$ cooperates together with the generative weights $\mathbf{W}_2$ from the second layer to construct the generative random Fourier feature in the second layer
$\mathbf{z}_i^2$ in the same way:
\begin{equation}
\mathbf{z}_i^2=
\phi(\mathbf{z}_i^1, \{\mathbf{w}_j^2\}^{D_2}_{j=1}), 
\mathbf{w}_j^2\sim \mathbb{P}_{k_2},
i=1,...,n.
\label{eq-GRFF-layer2}
\end{equation}
Again, notice that $\mathbf{w}^1_j$ and $\mathbf{x}_i$ are of the same dimension and that the dimension of $\mathbf{z}^1_i$ equals to $2D_1$.
Accordingly, the dimension of $\mathbf{w}^2_j$ equals to $2D_1$ and the dimension of $\mathbf{z}^2_i$ equals to $2D_2$.
Finally, there is a linear classifier $\Phi_C$ applied to the last random feature $\mathbf{z}_i^2$ to output the prediction.

Now we rewrite the ERM problem in Eq.\eqref{loss-EmpRM} for the two-layer GRFFs as follows,
\begin{equation}
\mathop{\min}_{\theta_{G_1},\theta_{G_2},\theta_C}
\frac{1}{n}
\sum_{i=1}^{n}
L\Big(
\Phi_C\big(
\phi\big(\mathbf{z}_i^1, \Phi_{G_2}(\mathbf{N}_2)\big)
\big)
, y_i\Big),\ 
\mathbf{z}_i^1 = \phi\big(\mathbf{x}_i, \Phi_{G_1}(\mathbf{N}_1)\big).
\label{loss-EmpRM-ml}
\end{equation}

\begin{figure}
	\centering
	\subfloat[Loss curves]{\includegraphics[scale=0.22]{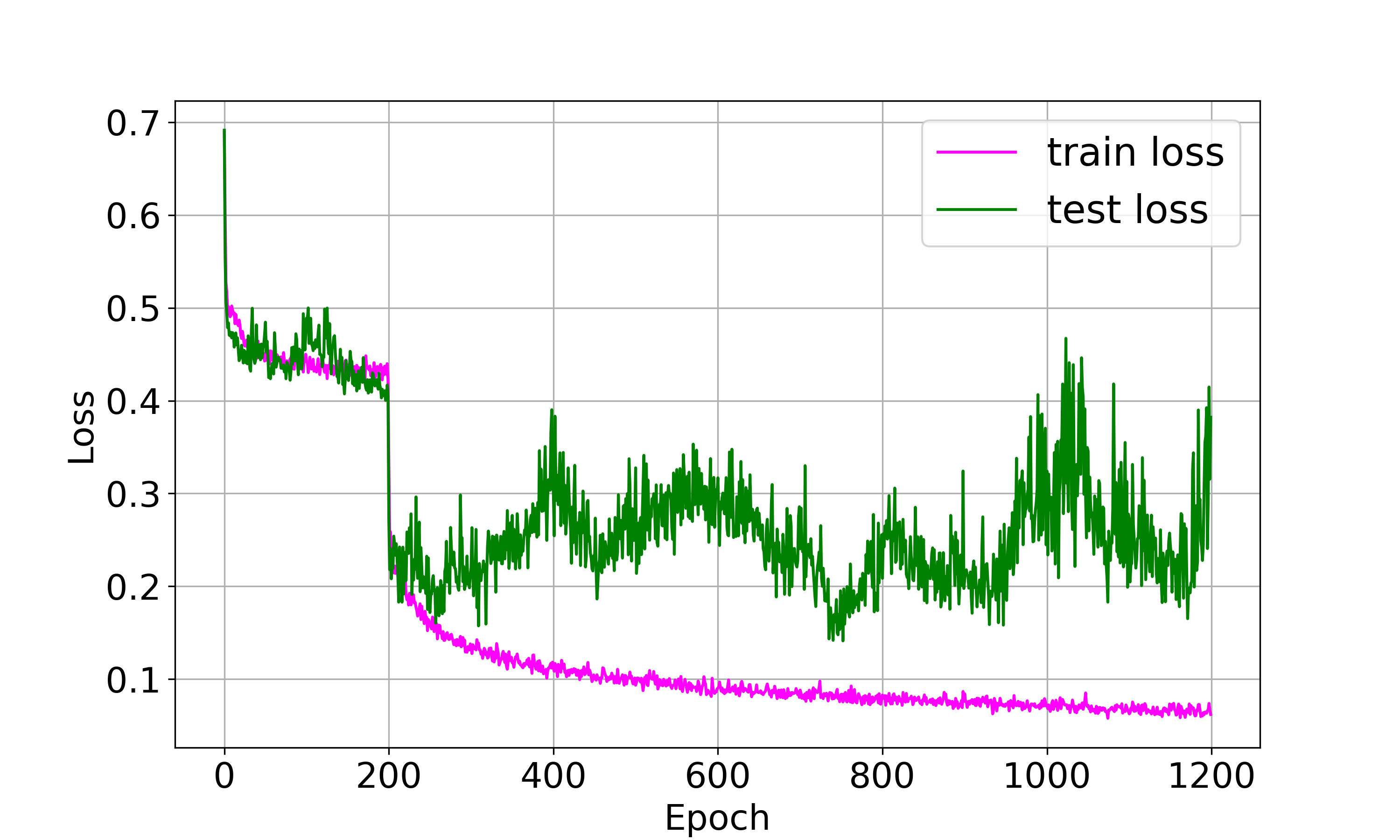}%
	}
	\hfil
	\subfloat[Accuracy curves]{\includegraphics[scale=0.22]{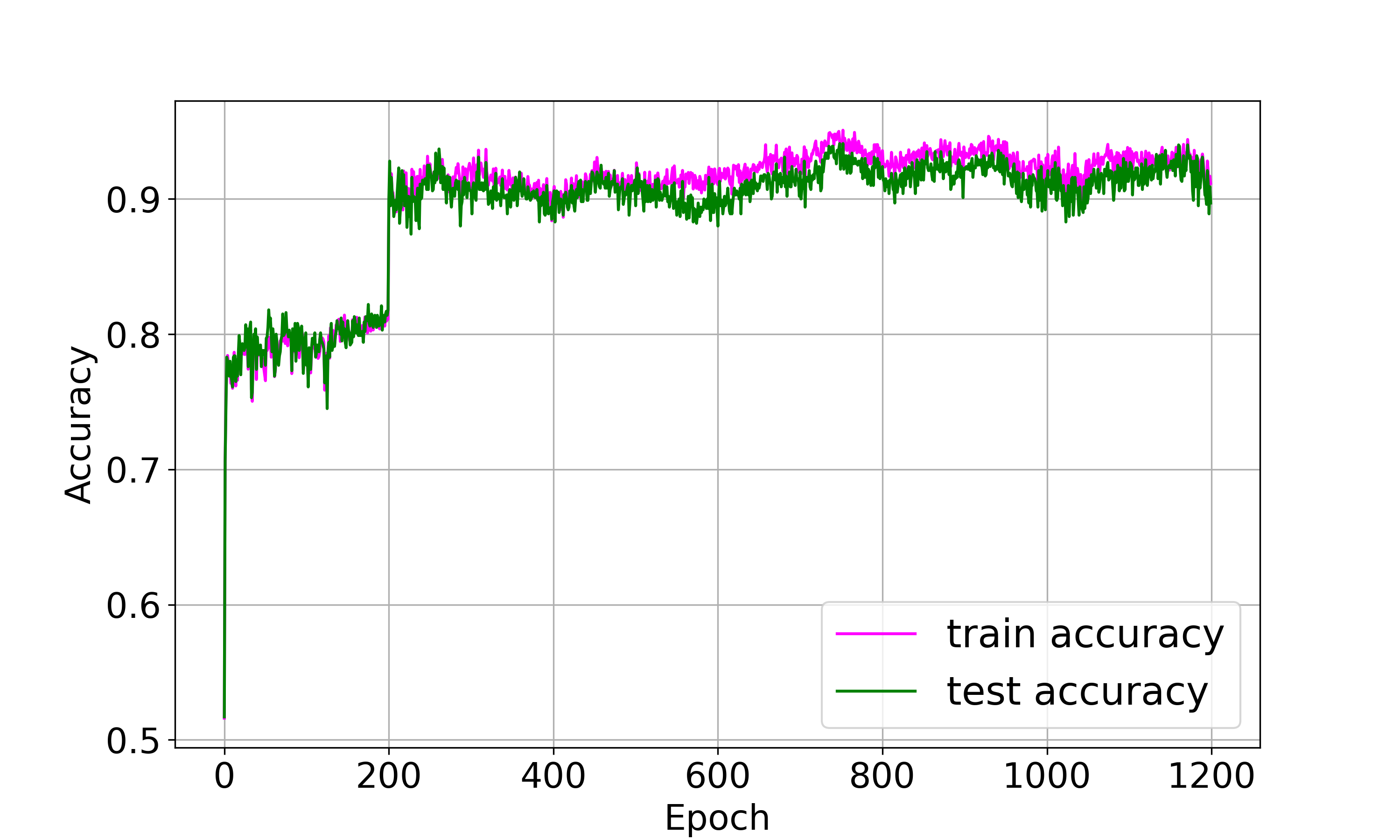}%
	}
	
	\caption{An example of the training process on a bi-classification task. A two-layer structure is adopted. Loss and accuracy variations are recorded. There are evident performance leaps, i.e., the drop of losses and the step-up of accuracies, by adding a second layer at the turning point of 200-th epoch. Detailed settings can be found in section \ref{sec-experimentsresults}.}
	\label{fig_GRFF_mulilayer_illustration}
\end{figure}

Advantages of the multi-layer structure are obvious.
The generator in the current layer actually models some distribution on features from the previous layer, indicating a good kernel on features.
Besides, associated with the proposed progressively training strategy (introduced in section \ref{subsec-train}), during the whole training process, by updating the generators layer-by-layer, there are distinct improvements on both the loss curves and the accuracy curves (see Fig.\ref{fig_GRFF_mulilayer_illustration}), implying that adding more layers leads to significant performance leaps.
More details about the distributions on features are illustrated in section \ref{sec-experimentsresults}.

\subsection{Progressively Training}
\label{subsec-train}

For the multi-layer generative random Fourier features,
since there exist several generative networks, it is hard to efficiently update all the parameters in multiple networks simultaneously, which possibly results in a total failure of convergence \cite{kawaguchi2019gradient}.
Therefore, inspired by the training in ProGAN \cite{karras2018progressive}, we propose a progressively training strategy to efficiently train the multiple generators in an inverse and layer-by-layer order.

The progressively training strategy contains several phases.
The number of phases equals to the number of generators.
In the first phase, we freeze the update of all the parameters except the last generator and the FC layer.
After the training converges, we step into the next phase.
Now the penultimate generator is unfrozen and gets updated together with the last one and the FC layer until convergence, while the others are still kept fixed.
In such an inverse and layer-by-layer order, we progressively unfreeze the generators in each layer, add them to the training sequence one by one, and finally train all these generators together with the FC layer,
which is empirically proved as an efficient training way for the multi-layer structure. 
% The training algorithm for a two-layer structure is demonstrated in Alg.\ref{alg-train}.
\fk{This progressively training alogrithm is demonstrated in Alg.\ref{alg-train}}.

In addition, more details of the generator are described.
Every generator is parameterized as a multi-layer perceptron (MLP), which can be viewed as a composition of several cascaded blocks.
Among these blocks, except for the last one, each block sequentially holds a linear layer, a batch normalization \cite{ioffe2015batch} layer, and an activation layer with leaky ReLU function.
In the last block, we remove the batch normalization layer and replace leaky ReLU with \emph{tanh} function as the activation.
Fig.\ref{fig_slgrff} illustrates the described components inside the generator.
Furthermore, one can customize the network according to the practical cases.
For example, it is allowed to increase the number of blocks in the MLP to enhance its learning ability or to add dropout \cite{srivastava2014dropout} to alleviate overfitting.
The number of neurons in the last FC layer can be modified freely for multi-classification or regression tasks.

\begin{algorithm}[!ht]
	\caption{\fk{The progressively training for GRFFs.}}
	\label{alg-train}
	\begin{algorithmic}[1]
		\REQUIRE \fk{Training set $X_{tr}$, number of generators $K$, numbers of epochs $epo_1,epo_2,\cdots,epo_K$ w.r.t. $K$ training phases.}
		\ENSURE \fk{$K$ generators $\Phi_{G_1},\Phi_{G_2},\cdots,\Phi_{G_K}$ and a linear classifier $\Phi_C$.}
		
		\WHILE{\fk{not reach $\sum_{i=1}^{K}epo_i$}}
		\STATE \fk{Sample $K$ sets of noises from $\mathbb{P}_0$: $\mathbf{N}_i,i=1,\cdots,K$.}
		\STATE \fk{Take a batch of samples $(\mathbf{x},y)$ from $X_{tr}$.}
		\STATE \fk{Build features at the $k$-th layer sequentially: \\
		$\mathbf{Z}^k=\phi(\mathbf{Z}^{k-1},\Phi_{G_k}(\mathbf{N}_k)),\ \mathbf{Z}^0=\mathbf{x},\ k=1,\cdots,K$.}
		\STATE \fk{Determine the loss: $loss=L(\Phi_C(\mathbf{Z}^K),y)$.}
        \FOR[\fk{the $j$-th training phase}] {\fk{$j=1,\cdots,K$}}
        \IF[\fk{inverse order}] {\fk{not reach $\sum_{i=1}^jepo_i$}}
		\STATE \fk{Update $\Phi_{G_{K-j+1}},\cdots,\Phi_{G_K}$ and $\Phi_{G_C}$ via $\nabla loss$ by Adam.}
		\STATE \fk{Break.}
		\ENDIF	
		\ENDFOR
		\ENDWHILE
	\end{algorithmic}
\end{algorithm}

The standard normal distribution is set as the input noise distribution $\mathbb{P}_0$, which corresponds to a radial basis function (RBF) kernel, the most widely used universal kernel.
At each iteration, we independently \emph{resample} noises from $\mathbb{P}_0$ to optimize the generators on the expectation of the distribution, and update \emph{simultaneously} the parameters of the generators and the classifier by Adam \cite{kingma2014adam} optimizer.
We choose the cross entropy function as the loss function $L$.
One can estimate whether the model is trained until convergence by watching the variation of the cross entropy loss value, with which the final classification performance is closely connected.

For the inference process, given a new sample $\mathbf{x}_{new}$, 
random noises are first sampled from $\mathbb{P}_0$, then the weights are generated via the generators, then the corresponding generative random Fourier feature $\mathbf{z}_{new}$ is constructed via the weights and $\mathbf{x}_{new}$ by applying Eq.\eqref{eq-GRFF} or Eq.\eqref{eq-GRFF-layer1}, Eq.\eqref{eq-GRFF-layer2}. Finally, the linear classifier $\Phi_C$ will predict the label of $\mathbf{x}_{new}$.

So far, we finish building an end-to-end, one-stage kernel learning method, which implicitly learns the kernel distribution by solving an ERM problem via the generative RFFs.
To learn the latent kernel distribution, 
a generative network is designed to simulate the sampling process without an explicit definition of the probability density function of the distribution.
Besides, jointly learning the features and the classifier allows us to increase the depth of features.
The resulting multi-layer structure can learn a good kernel on features and thus even boosts the generalization performance.
In addition, a progressively training strategy is proposed to efficiently train the multiple generative networks.

\noindent\textbf{Remark}\quad We would like to reiterate the following facts. 
It's worth noting that the initial motivation of RFFs-based methods is to reduce the time and space complexity in large-scale kernel machines. 
For instance, the vanilla RFFs \cite{rahimi2008random} speed up the kernel machines a lot, then get even accelerated in \cite{sinha2016learning}.
Further, random features are then extended to kernel learning, since RFFs actually provide a good justification for kernel learning in the spectral cases.
The researches in \cite{sinha2016learning,li2019implicit} are two typical kernel learning algorithms via random features, see a recent survey \cite{surveyRFF} on the development of RFFs-based algorithms for kernel learning.
Regarding to our method, the target is to perform kernel learning via RFFs, but NOT reducing the time-consuming.
In fact, since the neural network requires training, our method and the method in \cite{li2019implicit} are inevitably inferior in the processing speed during training and inference, compared with those algorithms specifically designed for kernel acceleration \cite{rahimi2008random,sinha2016learning}.
Instead, the purpose of our method is to learn a more well-generalized kernel based on the fact that the optimization objective in \cite{sinha2016learning,li2019implicit}, i.e., solving the target alignment, is not necessarily optimal in classification tasks.
To achieve such a purpose, we introduce the generative networks and adopt an end-to-end scheme to reach a one-stage solution.
The resulting two merits of this solution, better generalization and stronger robustness, will be empirically illustrated in the next section.

\section{Experiments and Results}
\label{sec-experimentsresults}
We evaluate the classification performance of the proposed GRFFs on a wide range of data sets.
Following the work in \cite{sinha2016learning,li2019implicit}, we first test the performance on a synthetic data set.
Second, we choose several benchmark data sets from UCI repository$\footnote{https://archive.ics.uci.edu/ml/datasets.html}$ and LIBSVM data$\footnote{https://www.csie.ntu.edu.tw/˜cjlin/libsvmtools/datasets/}$, including both large-scale and small-scale sets, and test the performance on these data sets.
Finally, we introduce a variant of the multi-layer GRFFs for image data and conduct an adversarial attack on this variant to illustrate its adversarial robustness.
% All the experiments are executed on a workstation with a single NVIDIA GPU GTX 1070.

The simplest version of GRFFs only includes a single layer, i.e., one single generative network, denoted as SL-GRFF.
We denote the multi-layer GRFFs including more than one generators as ML-GRFF.
We compare SL-GRFF and ML-GRFF with the following approaches.
\begin{itemize}
	\item RFF. \citet{rahimi2008random} first proposed the vanilla RFFs to approximate the kernel function by sampling from its spectral distribution. This approach is data-independent.
	\item OPT-RFF. Based on the vanilla RFFs, \citet{sinha2016learning} proposed to optimize the random features by solving a target alignment problem. A feature subset of an optimal size is learned, which shows great superiority in high processing speed in large-scale kernel machines.
	\item IKL. \citet{li2019implicit} adopted a neural network to model the spectral distribution of the kernel function. The network is trained by optimizing a target alignment problem. A linear classifier is applied on the features.
	\item MLP. The traditional MLP with ReLU activation function is also included in the comparison.
	Specifically, the numbers of layers in the MLP equal to that in the ML-GRFF, and the numbers of neurons in each hidden layer in the MLP equal to the numbers of sampled noises in each layer in the ML-GRFF.
\end{itemize}

\fk{In all the experiments, for RFF and OPT-RFF, the random features are generated by sampling weights from a normal distribution, which makes the features approximate an RBF kernel.
Regarding to the number of sampled weights, i.e., the value of $D$ in Eq.(\ref{eq-RFF}), following the implement\footnote{\href{https://github.com/duchi-lab/learning-kernels}{https://github.com/duchi-lab/learning-kernels}\label{footnote:codes}} of OPT-RFF \cite{sinha2016learning}, a large enough number of weights are initially sampled in OPT-RFF to learn a feature subset of an optimal size $D_{opt}$.
Then this learned $D_{opt}$ is employed to build the vanilla  RFFs  \cite{rahimi2008random}.}
The ridge regression is set as the linear classifier in the second stage of RFF and OPT-RFF.
We only compare with the results of IKL claimed in the paper \cite{li2019implicit} on the synthetic data set in section \ref{subsec-syn}.

For our proposed GRFFs, in section \ref{subsec-syn} and \ref{subsec-ben}, for the multi-layer version, we adopt a two-layer structure including two generators, each of which is parameterized as an MLP.
We set $D_1=256$ and $D_2=64$.
The structures of the first and second generators are $100\to 128 \to 64 \to 64 \to dim$ and $100\to 512 \to 256 \to 256 \to 512$ respectively, where $dim$ denotes the input data dimension.
The FC layer contains $2D_2=128$ neurons.
For the single-layer GRFFs in section \ref{subsec-syn},
the generator is also parameterized as an MLP and holds a structure of $100\to 128 \to 64 \to 64 \to dim$.
We set $D = 256$, thus the following FC layer contains $2D=512$ neurons.
\fk{We discuss GRFFs with more than 2 generators in \ref{sec:appendix}.}

\subsection{Performance on Synthetic Data}
\label{subsec-syn}
\begin{figure}[!t]
	\centering
	\subfloat[Synthetic data when $d=2$]{\includegraphics[scale=0.24]{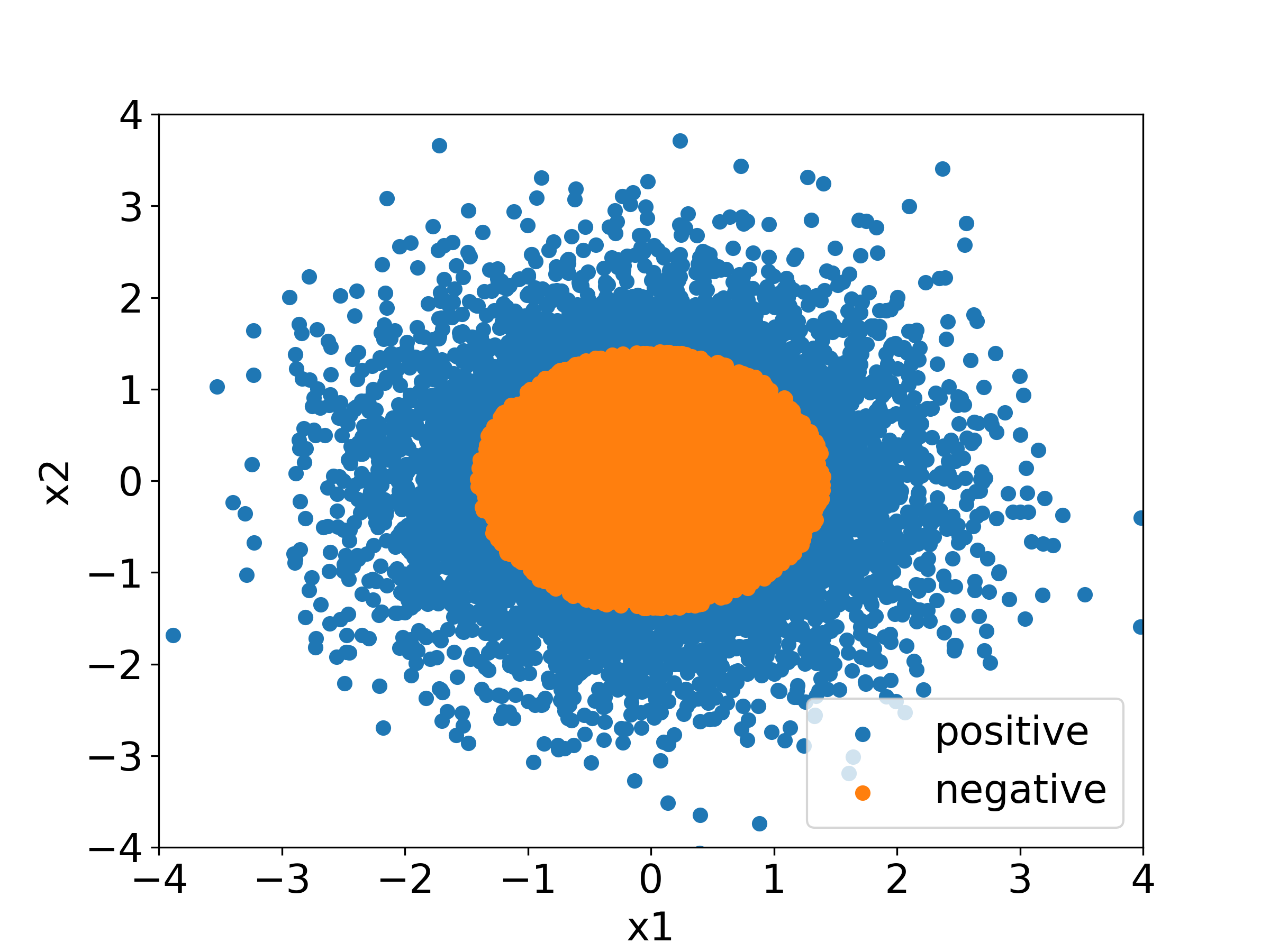}%
		\label{fig_syn_data}}
	\hfil
	\subfloat[Test errors vs. dimension]{\includegraphics[scale=0.24]{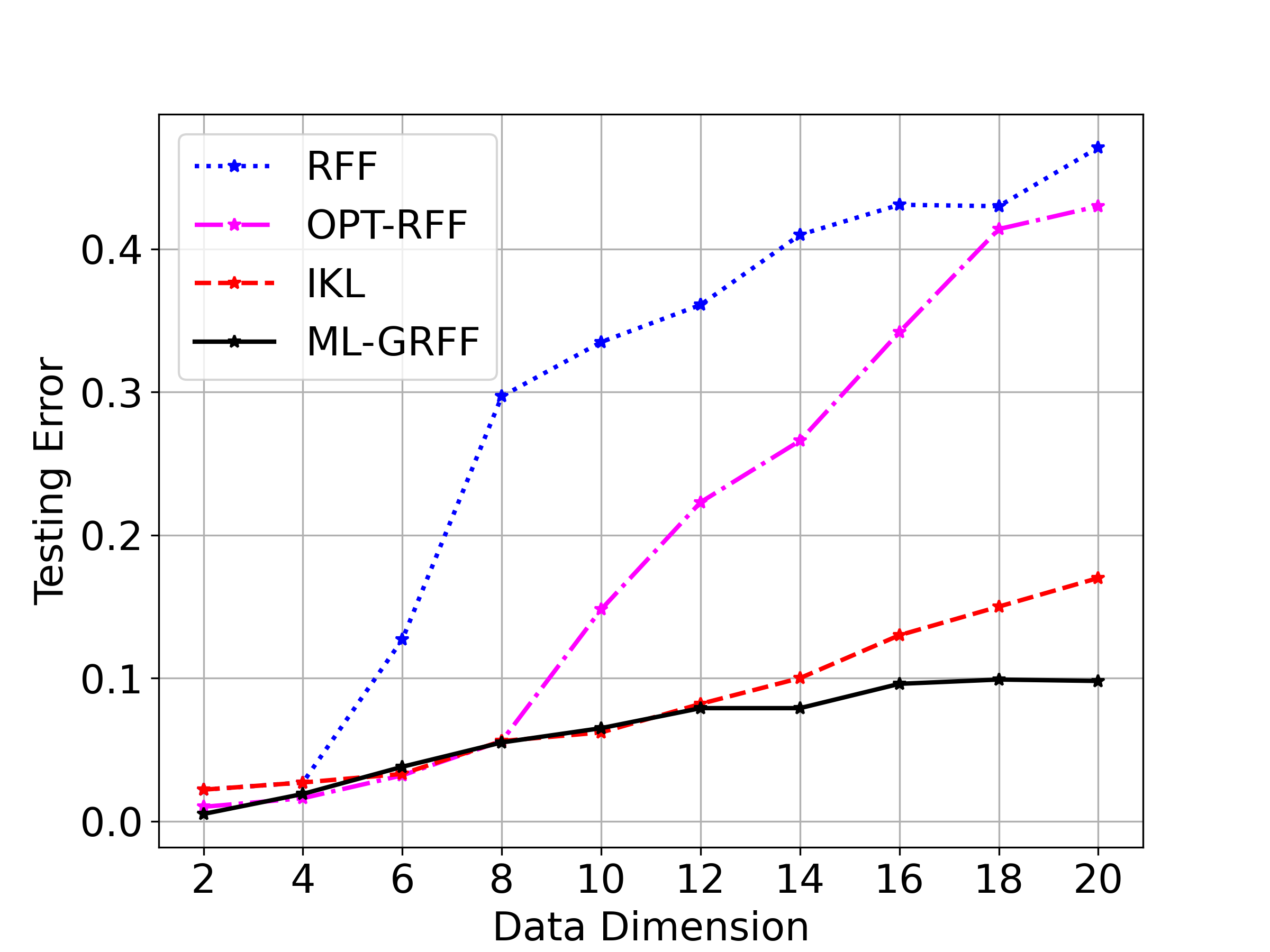}%
		\label{fig_syn_error1}}
	\hfil
	\subfloat[Errors vs. dimension]{\includegraphics[scale=0.24]{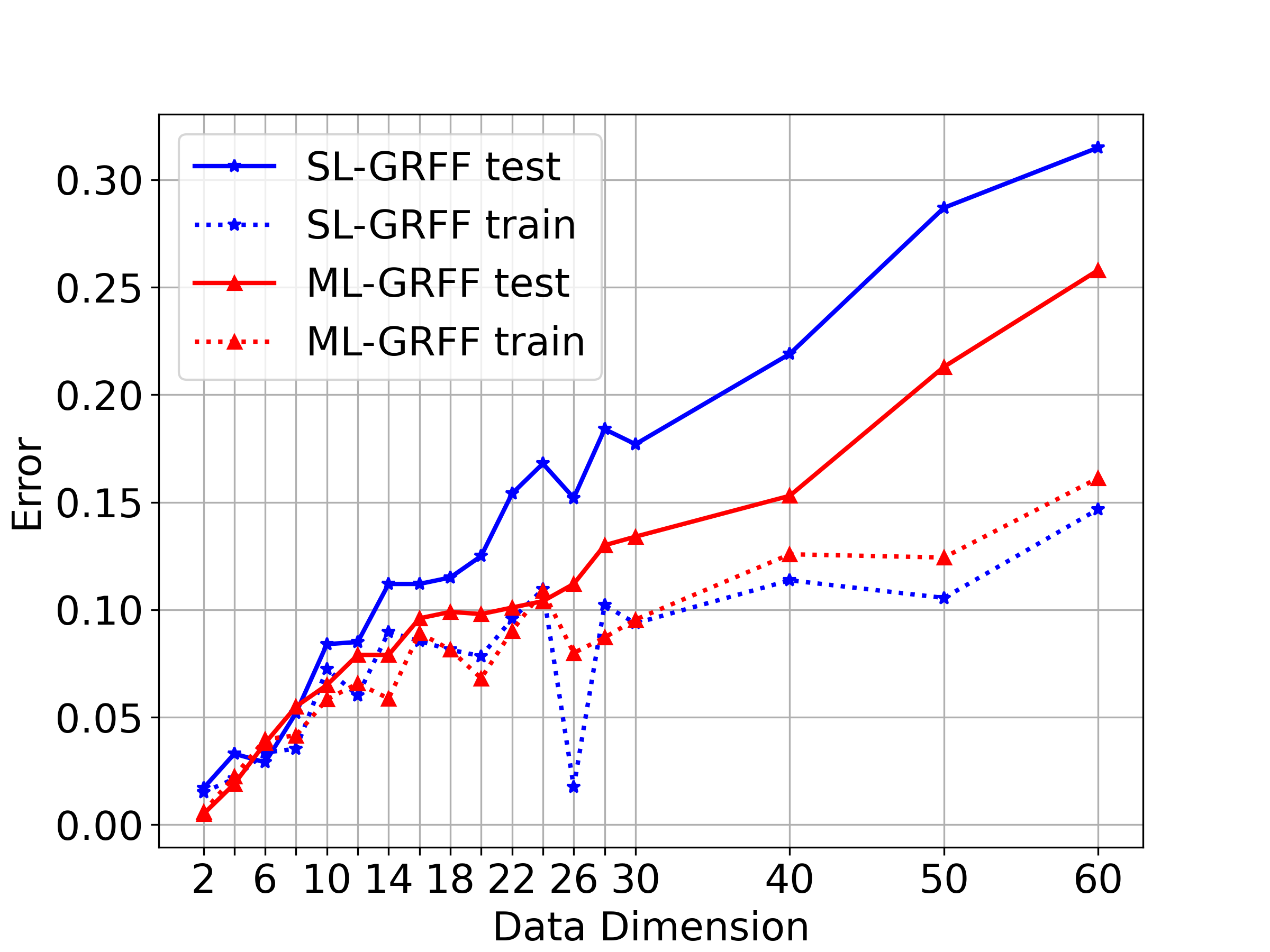}%
		\label{fig_syn_error2}}
	\caption{Performance on the synthetic data. (a) Data distribution when the dimension equals to 2. (b) Misclassification errors on the test set of different methods.
		(c) Misclassification errors on the training and test sets of SL-GRFF and ML-GRFF.}
	\label{fig_synthetic}
\end{figure}

For the synthetic set, we generate $\{\mathbf{x}_i\}_{i=1}^{n} \sim \mathcal{N}(0, I_d)$ with $y_i=\mathrm{sign}(\|\mathbf{x}_i\|^2-\sqrt{d})$, where $d$ is data dimension.
The training set includes ${10}^4$ samples and the test set contains ${10}^3$ samples.
Fig.\ref{fig_syn_data} shows the data distribution when $d=2$.
% One can find that the RBF kernel is ill-suited for this data set \cite{sinha2016learning}.
\fk{An RBF kernel is ill-suited for this data set since the Euclidean distance used in this kernel fails to correctly capture the data pattern near the boundary \cite{sinha2016learning}.
One may need to carefully tune the kernel width to alleviate the inherent inappropriateness between the RBF kernel and the data set.}

\begin{figure*}[h]
	\centering
	\subfloat[PCA on SL-GRFF, $d=8$.]{\includegraphics[scale=0.22]{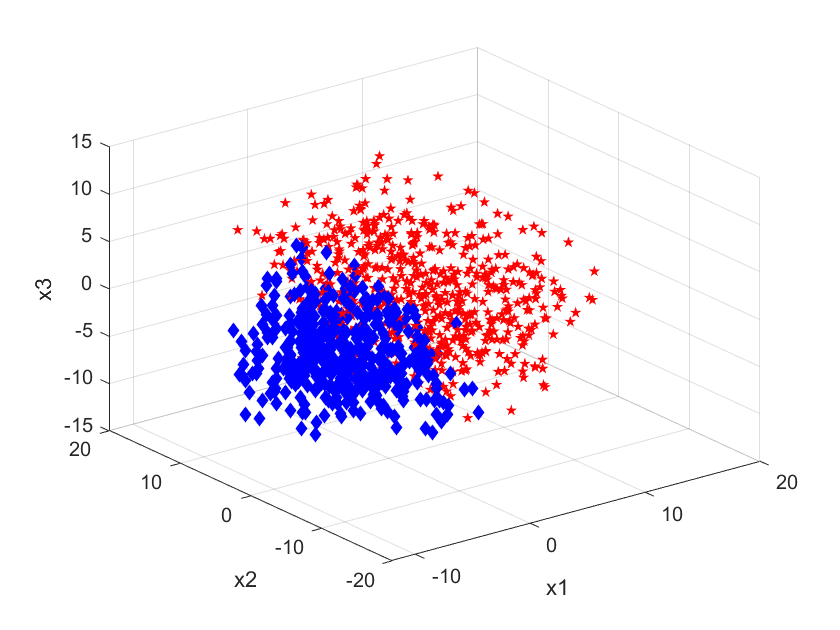}%
		\label{fig_syn_pca_SLGRFF_dim8}}
	\hfill
	\subfloat[PCA on ML-GRFF in the 1st layer, $d=8$.]{\includegraphics[scale=0.22]{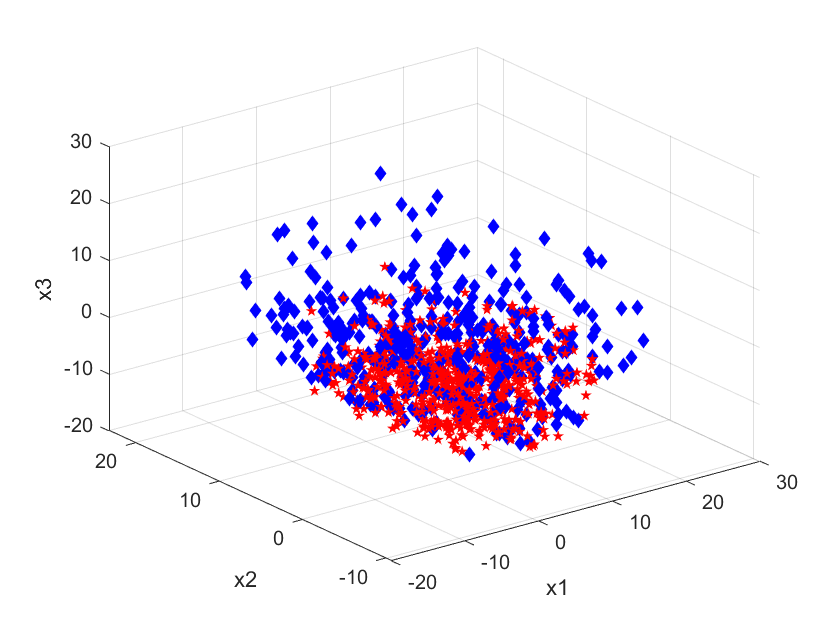}%
		\label{fig_syn_pca_MLGRFF_layer1_dim8}}
	\hfill
	\subfloat[PCA on ML-GRFF in the 2nd layer, $d=8$.]{\includegraphics[scale=0.22]{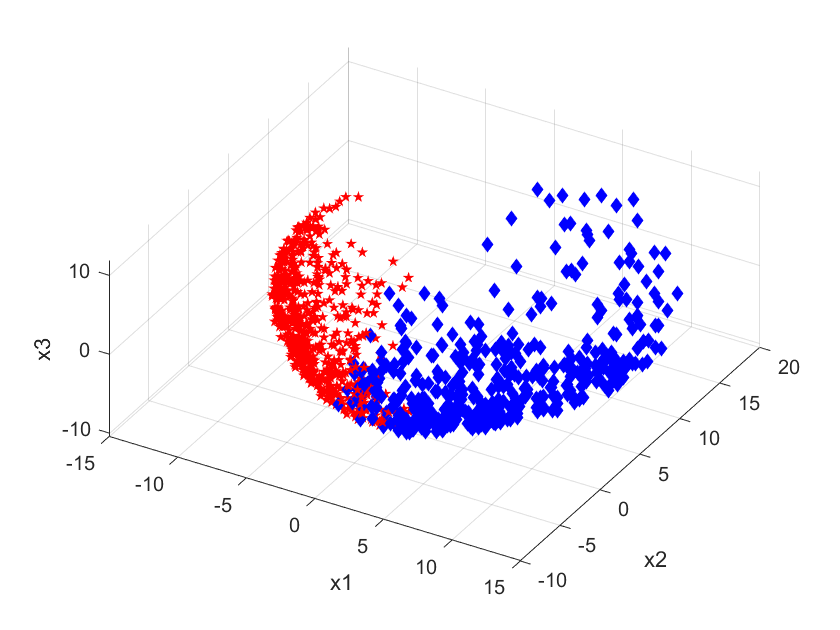}%
		\label{fig_syn_pca_MLGRFF_layer2_dim8}}
	
	\vfill
	
	\subfloat[PCA on SL-GRFF, $d=24$.]{\includegraphics[scale=0.22]{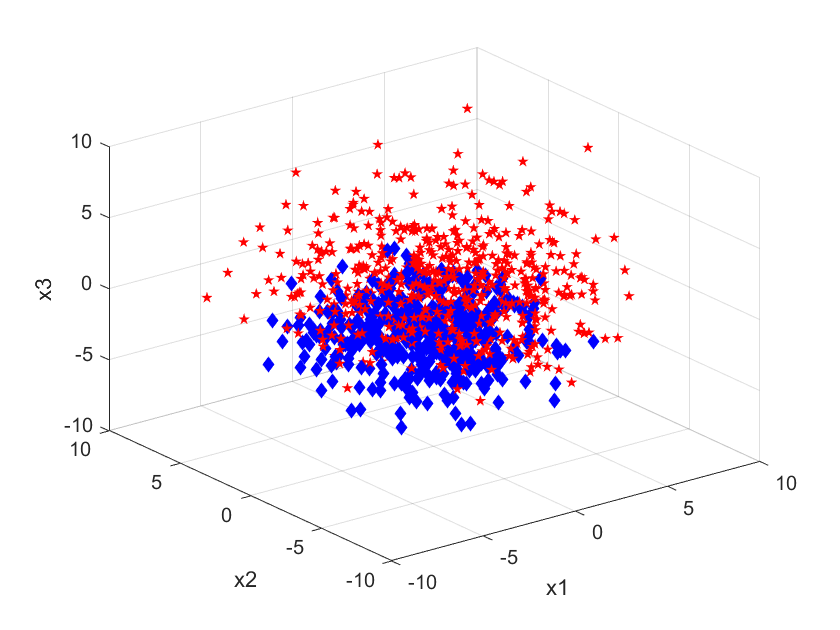}%
		\label{fig_syn_pca_SLGRFF_dim24}}
	\hfill
	\subfloat[PCA on ML-GRFF in the 1st layer, $d=24$.]{\includegraphics[scale=0.22]{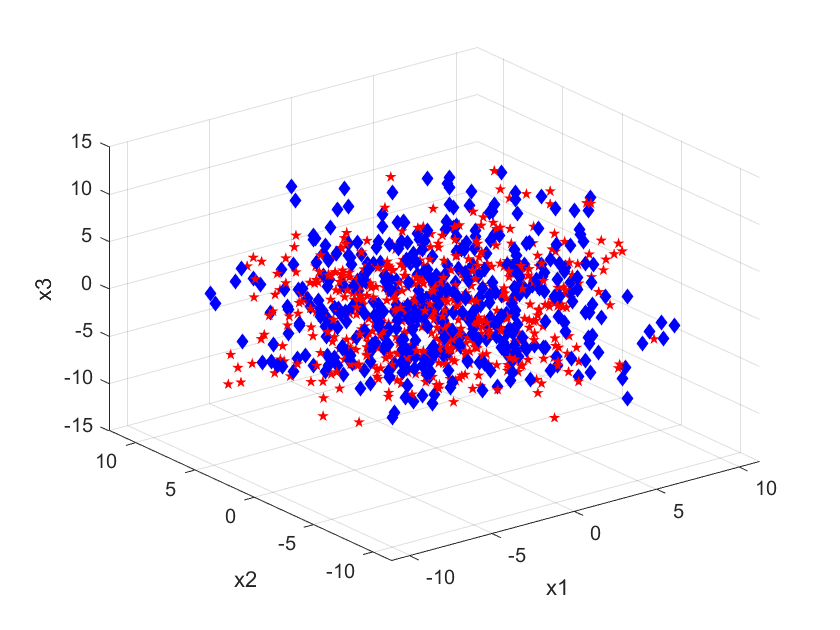}%
		\label{fig_syn_pca_MLGRFF_layer1_dim24}}
	\hfill
	\subfloat[PCA on ML-GRFF in the 2nd layer, $d=24$.]{\includegraphics[scale=0.22]{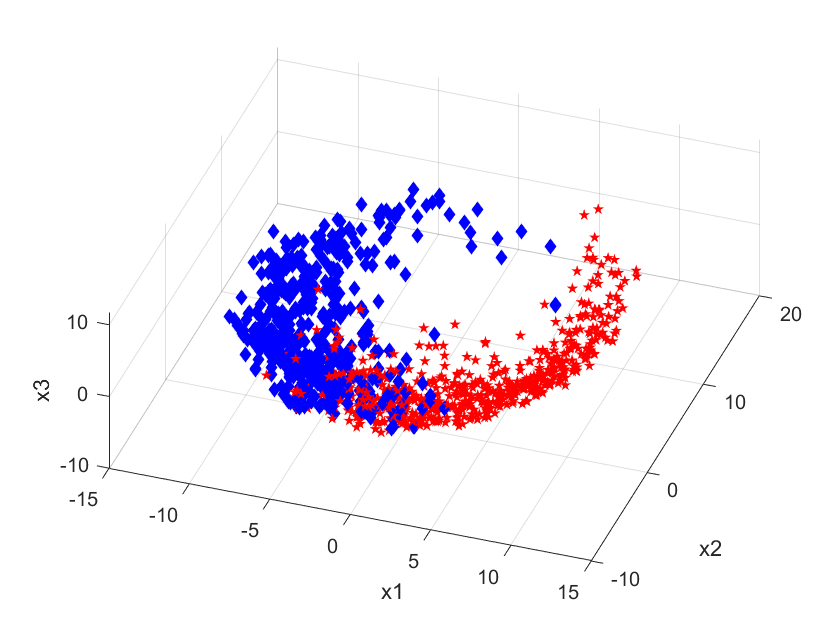}%
		\label{fig_syn_pca_MLGRFF_layer2_dim24}}	
	
	\caption{PCA visualizations of features of SL-GRFF and ML-GRFF, corresponding to synthetic data when $d=8$ (top) and $d=24$ (bottom) respectively. The blue diamond points denote positive samples, while the red pentagram points denote negative samples.}
	\label{fig_syn_pca}
\end{figure*}

\begin{figure*}[!t]
	\centering
	
	\subfloat[Loss curves when $d=2$]{\includegraphics[scale=0.3]{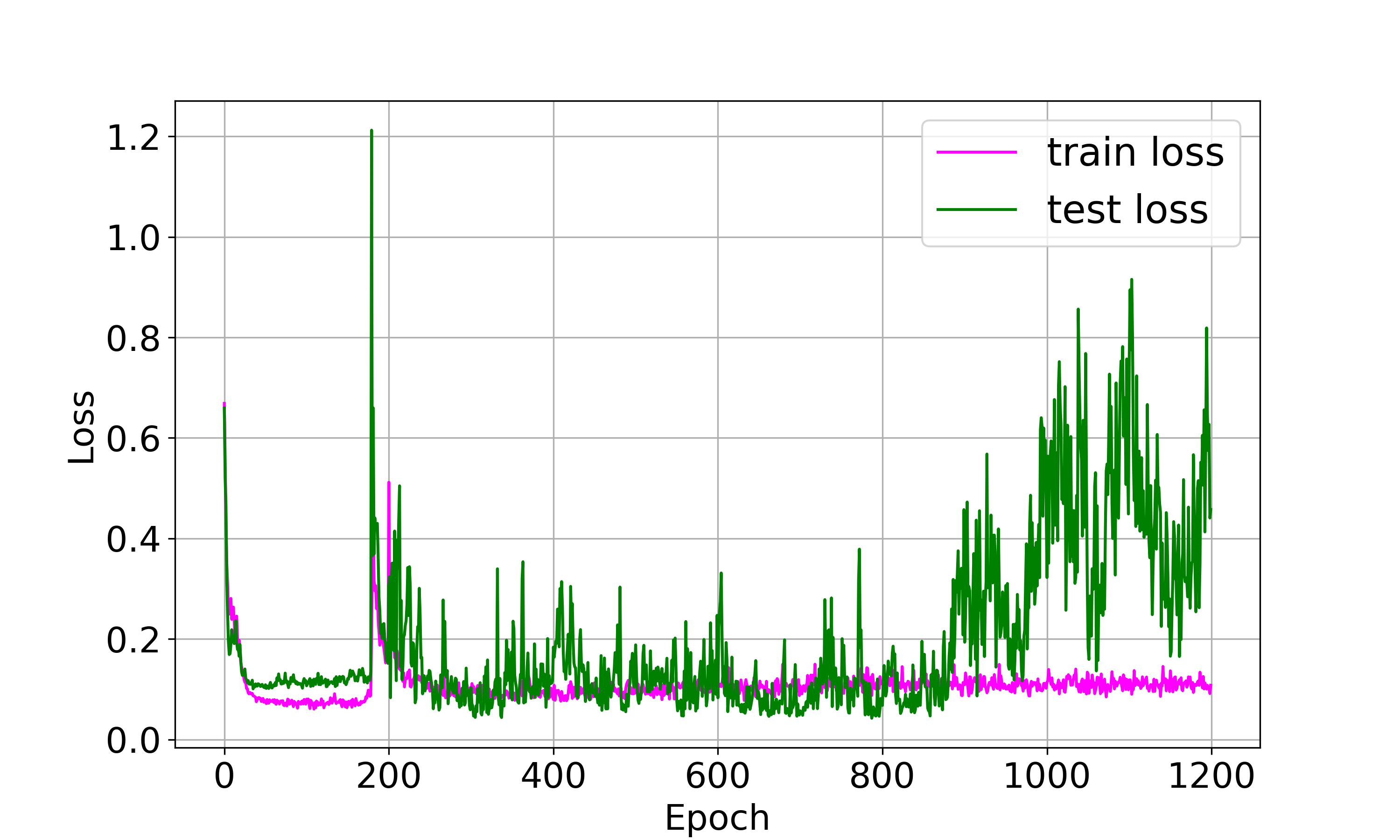}%
		\label{fig_syn_loss_dim2}}
	\hfill
	\subfloat[Accuracy curves when $d=2$]{\includegraphics[scale=0.3]{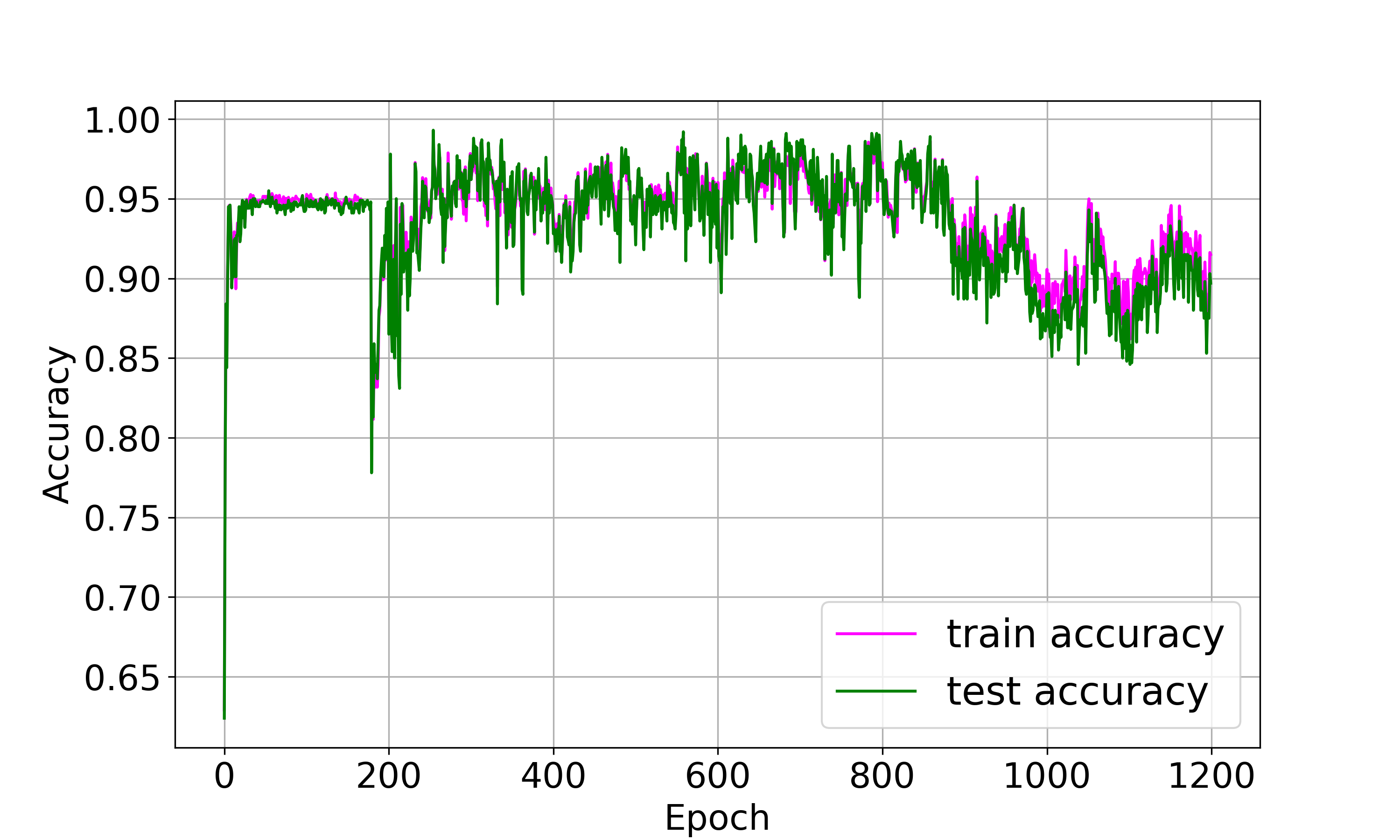}%
		\label{fig_syn_acc_dim2}}
	
	\vfill
	
	\subfloat[Loss curves when $d=6$]{\includegraphics[scale=0.3]{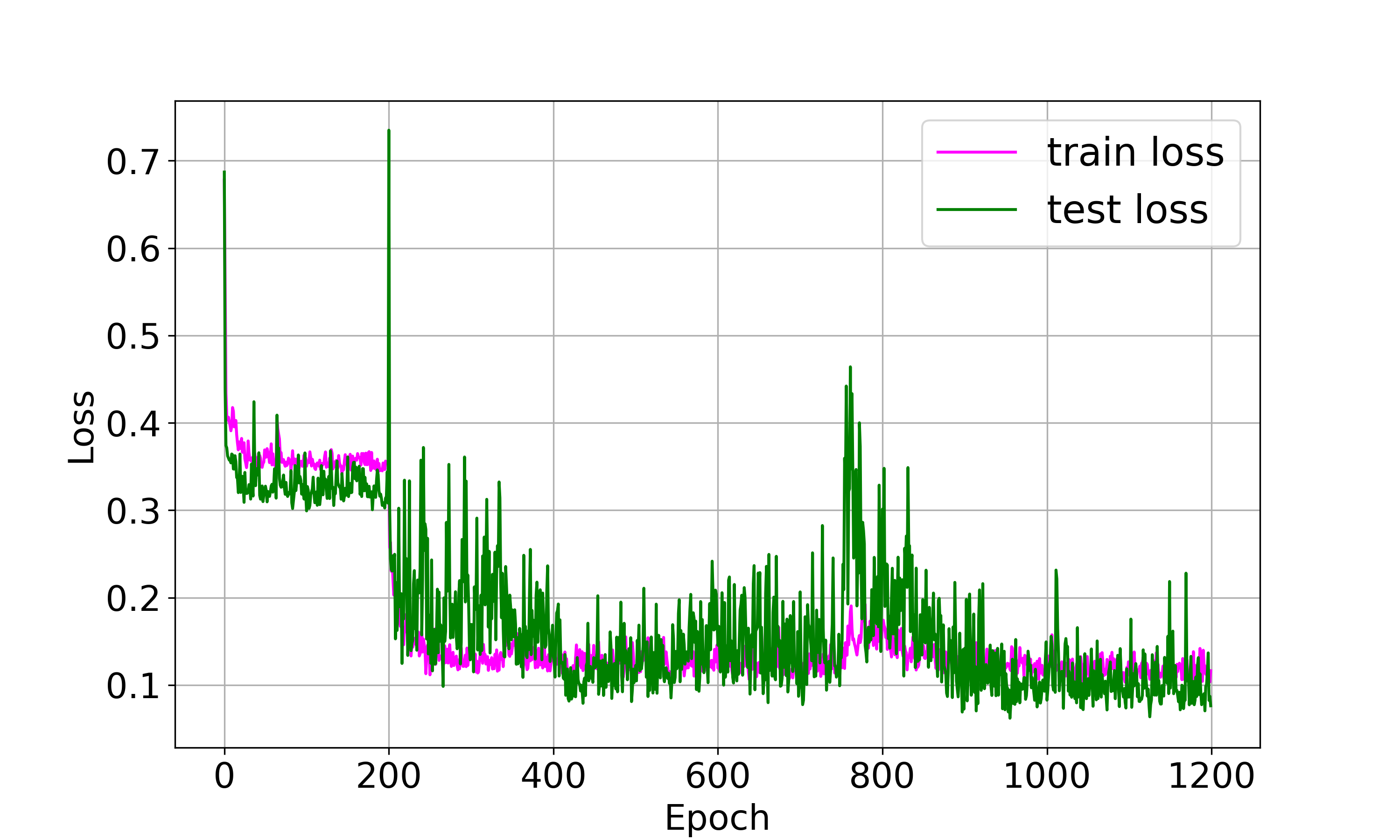}%
		\label{fig_syn_loss_dim6}}
	\hfill
	\subfloat[Accuracy curves when $d=6$]{\includegraphics[scale=0.3]{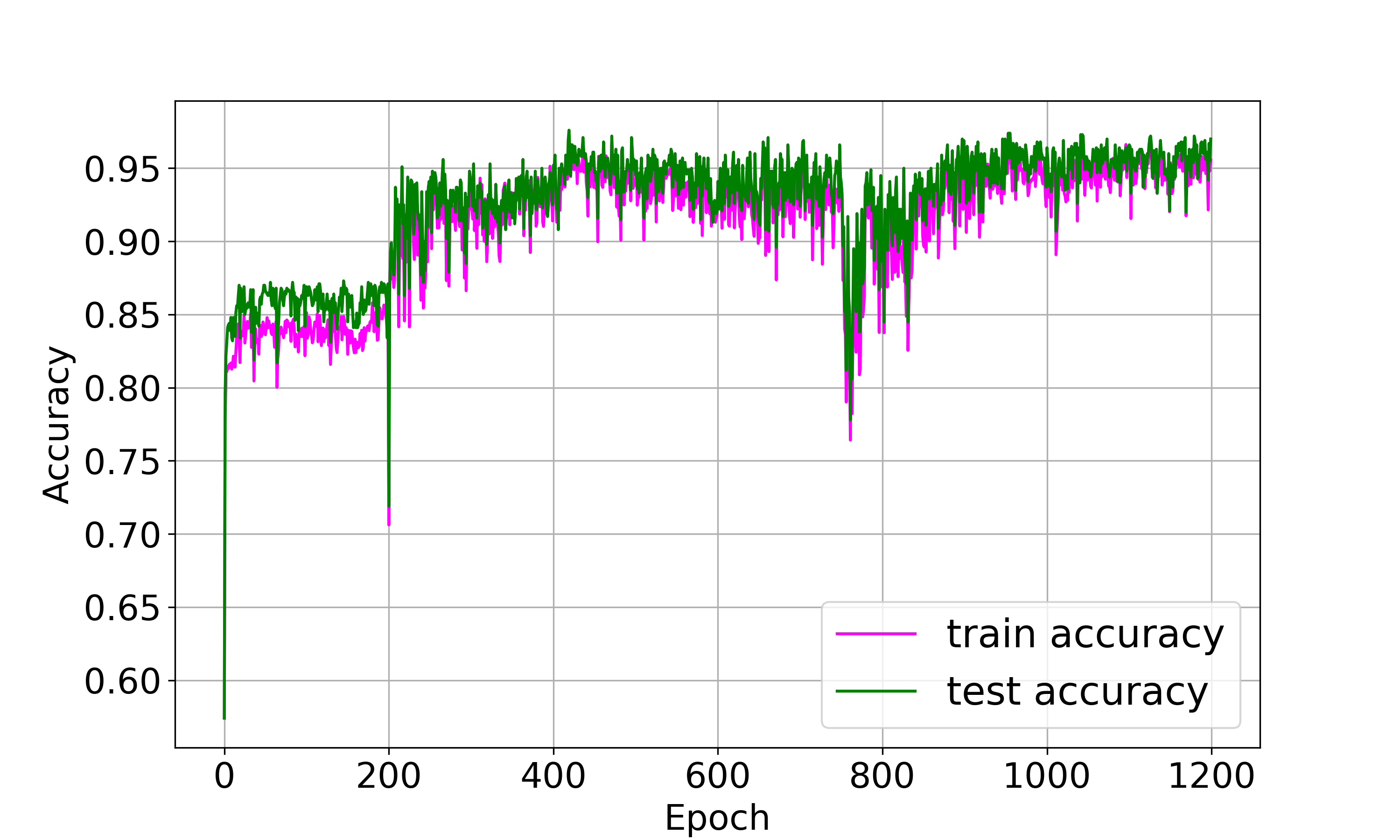}%
		\label{fig_syn_acc_dim6}}
	
	\vfill	
	
	\subfloat[Loss curves when $d=8$]{\includegraphics[scale=0.3]{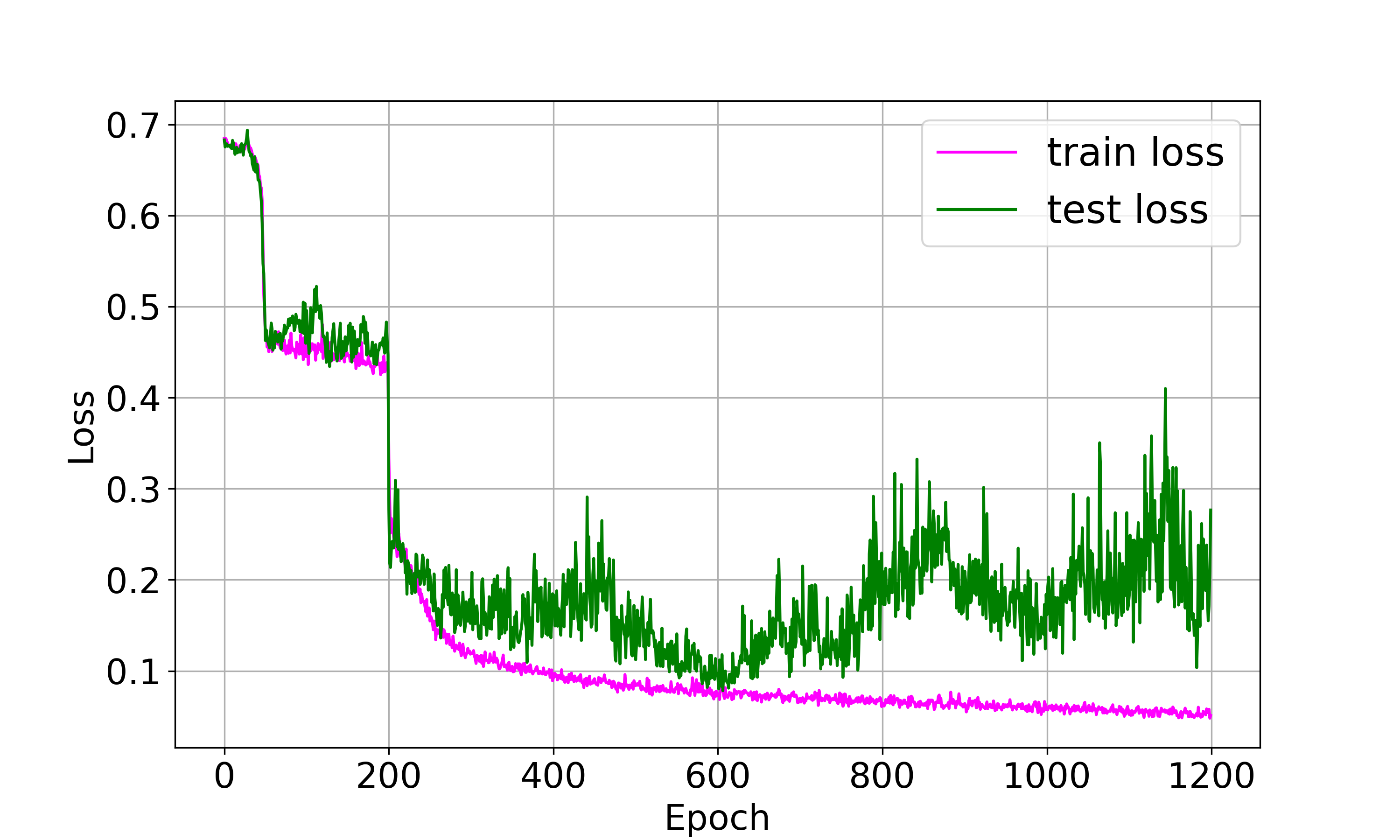}%
		\label{fig_syn_loss_dim8}}
	\hfill
	\subfloat[Accuracy curves when $d=8$]{\includegraphics[scale=0.3]{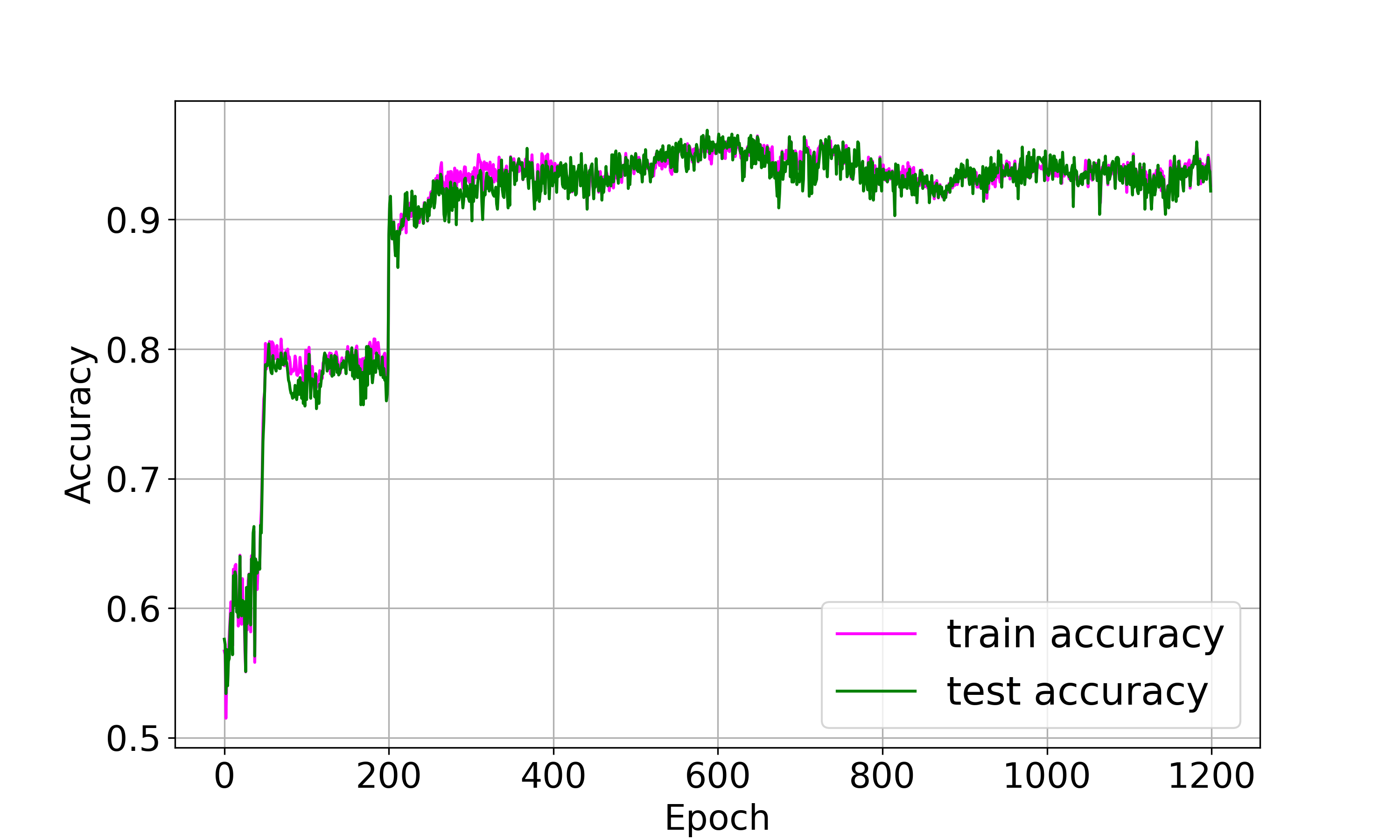}%
		\label{fig_syn_acc_dim8}}
	
	%	\vfill	
	%	
	%	\subfloat[Loss curves when $d=10$]{\includegraphics[scale=0.19]{fig/exp/losses-dim-10.png}%
	%		\label{fig_syn_loss_dim10}}	
	%	\hfill
	%	\subfloat[Accuracy curves when $d=10$]{\includegraphics[scale=0.19]{fig/exp/accuracies-dim-10.png}%
	%		\label{fig_syn_acc_dim10}}	

	\caption{Average loss and accuracy at each epoch on the synthetic data of different dimensions. The green curves denote records on the test set, while the magenta curves denote records on the training set. The turning point of the 1st and 2nd stages is the 200-th epoch.}
	\label{fig_syn_loss_acc}
\end{figure*}

Fig.\ref{fig_syn_error1} illustrates the test errors of different methods corresponding to different data dimensions $d\in\{2,4,6,8,...,18,20\}$.
Both RFF and OPT-RFF manifest a sharp performance decrease along with the increase of $d$.
By contrast, IKL and ML-GRFF both have rather stable performance as the dimension $d$ increases, since they try to learn the kernel distribution via a neural network without any prior kernel function definition.
Particularly, by directly optimizing the ERM problem in an end-to-end manner, ML-GRFF characterizes the data pattern much better than the others and achieves the lowest test errors.
Fig.\ref{fig_syn_error2} shows the comparison results on the training and test sets of SL-GRFF and ML-GRFF respectively.
In a wide range of data dimensions,
ML-GRFF always has an evident superiority of classification errors on the test set over SL-GRFF.

Besides, to visualize the generative RFFs, we take PCA \cite{wold1987principal} to extract the top-3 principal components of random features in SL-GRFF and different layers of ML-GRFF, shown in Fig.\ref{fig_syn_pca}.
SL-GRFF and ML-GRFF are trained on the synthetic data of two dimensions, $d=8$ and $d=24$, respectively.
When $d=8$, both SL-GRFF and ML-GRFF have similar performance, while when $d=24$, ML-GRFF performs better than SL-GRFF.
For ML-GRFF, the extracted principal components in the 1st layer (see Fig.\ref{fig_syn_pca_MLGRFF_layer1_dim8} and Fig.\ref{fig_syn_pca_MLGRFF_layer1_dim24}) do not show any evident linear separability.
But, the components in the 2nd layer (see Fig.\ref{fig_syn_pca_MLGRFF_layer2_dim8} and Fig.\ref{fig_syn_pca_MLGRFF_layer2_dim24}) follow a clear linear separable distribution.
For SL-GRFF, we can also see the linear separability of the extracted components from Fig.\ref{fig_syn_pca_SLGRFF_dim8} and Fig.\ref{fig_syn_pca_SLGRFF_dim24}, which is obviously weaker than that of the extracted components of the 2nd layer in ML-GRFF.
Hence, by going deeper and deeper, ML-GRFF is able to learn a good kernel on features, which results in better performance.

To illustrate the traits of the progressively training strategy, we record the loss and accuracy curves during training on the synthetic data of different dimensions in Fig.\ref{fig_syn_loss_acc}.
When the data is easy to be separated, i.e., it is low-dimensional ($d=2$), the whole training process does not show much differences between the two training phases in Fig.\ref{fig_syn_loss_dim2} and Fig.\ref{fig_syn_acc_dim2}.
As the dimension increases, there are obvious improvements, i.e., the drop of loss and the step-up of accuracy, from the 1st training phase to the 2nd training phase in Fig.\ref{fig_syn_loss_dim6} $\sim$ Fig.\ref{fig_syn_acc_dim8}, at the turning point of the 200-th epoch.
By progressively training the generators layer by layer, the convergence is achieved and there is good performance.

\subsection{Performance on Benchmark Data Sets}
\label{subsec-ben}

The models are trained on $7$ small-scale data sets and $3$ large-scale data sets.
We randomly pick half of the data as the training set, the other half as the test set, except for the data sets of which the training and test sets have already been divided.
After normalizing the data to ${[0,1]}^d$ in advance by min-max normalization, $20\%$ of the training data are randomly selected as the validation set.
\fk{For RFF and OPT-RFF, the validation set is adopted to perform validations on a group of 10 $\gamma$-s ranging from $0.5$ to $1.4$ with an interval of $0.1$, where $\gamma$ is the steep hyper-parameter in the RBF kernel: $k(\mathbf{x},\mathbf{x}')=\exp(-\gamma||\mathbf{x}-\mathbf{x}'||^2)$.
For MLP and ML-GRFF, during training, models that achieve best performance on the partitioned validation set are selected.}
All the experiments on every data set are repeated 5 times.

\begin{table}[t]
	\centering
	\caption{Training and test accuracy (\%) on small-scale data sets of different models.
		The highest test accuracy is highlighted in \textbf{bold}.}
	\label{tb-benchmark-smallscale}
	\resizebox{\textwidth}{!}{
		\begin{tabular}{c|c|ccccccc}
			\toprule[2pt]
			\multirow{2}{*}{method}  & {data set}
			& monks1 	& monks2 	& monks3 
			& australia & climate 	& diabetic & sonar \\
			& (\#tr;\#te;$d$)
			& (124;432;6)  & (169;432;6)  & (122;432;6) 
			& (345;345;14) & (270;270;18) & (576;575;19) & (104;104;60) \\
			\midrule[0.5pt]
			\multirow{2}{*}{RFF} 
			& train 
			& $95.48\pm3.15$ 	
			& $98.82\pm1.73$ 
			& $97.38\pm0.69$   
			& $93.51\pm2.57$ 	
			& $99.93\pm0.17$	
			& $79.90\pm2.00$ 
			& $100.00\pm0.01$ \\
			& test  
			& $81.48\pm1.83$ 	
			& $78.66\pm1.56$ 	
			& $93.61\pm0.45$   
			& $84.70\pm1.53$ 	
			& $91.48\pm1.72$ 	
			& $\mathbf{73.37\pm1.39}$ 
			& $77.88\pm4.90$ \\
			\midrule[0.5pt]
			\multirow{2}{*}{OPT-RFF} 
			& train 
			& $100.00\pm0.01$ 	
			& $91.83\pm3.28$ 	
			& $96.23\pm1.60$   
			& $93.45\pm2.29$ 	
			& $99.70\pm0.31$ 	
			& $78.78\pm2.08$ 
			& $100.00\pm0.01$ \\ 
			& test  
			& $91.67\pm4.29$ 	
			& $77.45\pm2.78$ 	
			& $92.13\pm1.96$   
			& $83.83\pm1.85$ 	
			& $92.44\pm1.42$ 	
			& $72.40\pm1.23$ 
			& $78.46\pm7.02$ \\
			\midrule[0.5pt]
			\multirow{2}{*}{MLP} 
			& train 
			& $96.94\pm0.32$ 	
			& $96.80\pm1.57$ 	
			& $97.05\pm0.40$   
			& $92.93\pm2.27$ 	
			& $98.37\pm0.83$ 	
			& $84.27\pm3.94$ 
			& $91.35\pm5.09$ \\ 
			& test  
			& $84.31\pm1.50$ 	
			& $\mathbf{84.86\pm2.98}$ 	
			& $85.37\pm1.46$   
			& $84.64\pm1.54$ 	
			& $93.26\pm1.13$ 	
			& $67.97\pm1.36$ 
			& $75.77\pm2.68$ \\		
			\midrule[0.5pt]
			\multirow{2}{*}{ML-GRFF} 
			& train 
			& $97.90\pm1.50$ 	
			& $82.72\pm1.96$ 	
			& $95.08\pm1.56$   
			& $85.86\pm1.93$ 	
			& $95.04\pm1.58$ 	
			& $69.53\pm2.01$ 
			& $80.38\pm7.08$ \\ 
			& test  
			& $\mathbf{95.83\pm2.15}$ 	
			& $79.72\pm1.18$ 	
			& $\mathbf{93.70\pm0.99}$   
			& $\mathbf{86.09\pm1.79}$ 	
			& $\mathbf{94.59\pm1.42}$ 	
			& $70.76\pm1.03$ 
			& $\mathbf{80.58\pm7.02}$ \\
			\bottomrule[2pt]
	\end{tabular}
	}
\end{table}

Tab.\ref{tb-benchmark-smallscale} and Tab.\ref{tb-benchmark-largescale} illustrate the training and test accuracy of different methods on the small-scale and large-scale data sets respectively.
The sizes of the training and test sets and the data dimensions are also listed in the two tables.
One can find that in most cases, ML-GRFF outperforms RFF and OPT-RFF a lot and achieves competitive (sometimes even slightly better) results compared with MLP.

\begin{table*}[t]
	\centering
	\caption{Training and test accuracy (\%) on large-scale data sets of different models.
		The highest test accuracy is highlighted in \textbf{bold}.}
	\label{tb-benchmark-largescale}
	% \resizebox{\textwidth}{!}{
	\begin{tabular}{c|c|ccc}
		\toprule[2pt]
		\multirow{2}{*}{method}  & {data set}
		& adult		     & ijcnn		& phishing \\
		& (\#tr; \#te; $d$)
		& (32,561; 16,281; 123)  & (49,990; 91,701; 22)  & (5,528; 5,527; 68) \\
		\midrule[0.5pt]
		\multirow{2}{*}{RFF} 
		& train 
		& $84.50\pm0.21$ 	
		& $95.35\pm0.19$ 	
		& $94.53\pm0.35$  \\
		& test  & 
		$84.25\pm0.16$ 	& 
		$93.18\pm0.19$ 	& 
		$92.63\pm0.61$  \\
		\midrule[0.5pt]
		\multirow{2}{*}{OPT-RFF} 
		& train 
		& $84.56\pm0.29$ 	
		& $94.85\pm0.28$ 	
		& $95.50\pm0.14$  \\
		& test  & 
		$84.47\pm0.28$ 	& 
		$93.20\pm0.05$ 	& 
		$94.26\pm0.22$  \\
		\midrule[0.5pt]
		\multirow{2}{*}{MLP} 
		& train 
		& $85.05\pm0.07$ 	
		& $98.58\pm0.48$ 	
		& $97.76\pm0.50$  \\
		& test  & 
		$84.70\pm0.19$ 	& 
		$\mathbf{98.54\pm0.23}$ 	& 
		$\mathbf{95.69\pm0.23}$  \\		
		\midrule[0.5pt]
		\multirow{2}{*}{ML-GRFF} 
		& train 
		& $84.93\pm0.14$ 	
		& $94.03\pm0.42$ 	
		& $95.23\pm0.32$  \\
		& test  & 
		$\mathbf{84.76\pm0.17}$ 	& 
		$94.60\pm0.57$ 	& 
		$95.20\pm0.38$  \\
		\bottomrule[2pt]
	\end{tabular}
	%}
\end{table*}

\subsection{\fk{Performance on Image Data and Adversarial Robustness}}
\label{subsec-imaa}
\fk{As aforementioned, the promising performance on generalization of the proposed one-stage, end-to-end kernel learning method is empirically validated on both synthetic and real-world data.
In the following, we conduct another experiment on adversarial robustness to demonstrate the superiority of our GRFF.}

\subsubsection{\fk{Background and Motivation}}

\fk{
The robustness of DNNs is demonstrated to be vulnerable and sensitive to imperceptible perturbations, a.k.a. adversarial examples \cite{szegedy2013intriguing,goodfellow2014explaining}, created by white-box adversarial attackers, which need availability to the complete information of DNNs, e.g., model parameters \cite{goodfellow2014explaining,kurakin2016adversarial}.
Hence, a line of adversarial defenses are based on an observation that adversarial examples created on one model by these attacks might totally lose efficacy on another one of different parameters \cite{fang2020towards}, which inspires us to investigate the adversarial robustness of GRFF.
}

\fk{
In GRFF, the independent samplings from the generators yield non-deterministic weights from the learned kernel distributions.
In other words, every time the independent sampling produces a GRFF model of different parameters.
Intuitively, such non-deterministic weights of GRFF raise the possibility of circumventing adversarial attacks on fixed parameters, thus bringing stronger adversarial robustness.
In the remainder of this subsection, we first describe a variant of GRFF for image data, then evaluate its adversarial robustness against an iterative, gradient-based adversarial attack  \cite{kurakin2016adversarial}.}

\subsubsection{\fk{Variant of GRFF for Image Data}}

\fk{
When dealing with image data, an image of $C\times H \times W$ size, where $C$,$H$,$W$ denote the number of channels, height and width respectively, can be simply stretched to a 1-dimensional vector. Then multiple 1-dimensional weight vectors of $C\times H \times W$ size could be generated to build the corresponding GRFFs of the image vector.
However, it is impractical to directly include so many $C\times H \times W$ neurons in the output layer of the generator, resulting in extremely numerous parameters and heavy computational burden.
Therefore, we design a variant of GRFF to deal with image data.
In the following, we elaborate two main changes of this variant in detail: how to generate weights and how to build random features.}

\fk{
To avoid posing $C\times H \times W$ neurons at the output layer, the generators are now devised to generate small-size, 2-dimensional weights, which resemble the common convolution kernels and could efficiently reduce computational costs.
Accordingly, the generator structure should be modified.
Specifically, the generators are parameterized the same as those in DCGAN \cite{radford2015unsupervised}, instead of MLPs.
Compared with the cascaded blocks in MLPs described in section 3.3 before, all the linear layers in the blocks are now substituted by transposed convolution \cite{zeiler2010deconvolutional} layers to generate 2-dimensional weights, and the leaky ReLUs are replaced by ReLUs.
The others still remain the same.}

\fk{
The way to build random features for image data still basically follows the feature map $\phi$ of Eq.(\ref{eq-RFF}). 
The inner product $\mathbf{w}^T\mathbf{x}$ between weights $\mathbf{w}$ and vectors $\mathbf{x}$ is now replaced by the convolution on image via the generated 2-dimensional weights.
The convoluted images then pass through cosine and sine functions respectively as indicated in Eq.(\ref{eq-RFF}), and the resulting two pieces are concatenated along the image channels, followed by a pooling operation.
These manipulations could also be sequentially executed on the output of a preceding layer to build random features at the current layer, thus forming a multi-layer structure.
After such multi-layer operations, i.e., convolution, cosine and sine activation, concatenation and pooling, the last random features are stretched to a 1-dimensional vector and an FC layer will make the predictions.}

\subsubsection{\fk{Performance against Adversarial Attacks}}

\fk{
Omitting the generators in the variant of GRFF described above, the model itself will reduce to a Convolution Neural Network (CNN) with cosine and sine activation functions.
The generators actually learn the convolution kernels in every layer of the CNN.
During inference, by resampling from the learned distributions for many times, the generators will keep generating non-deterministic and independent convolution kernels, which every time will lead to a specific CNN.
Hence, inspired by the randomized resampling mechanism associated with the distribution learning ability of GRFF, we investigate its application in defending the adversarial attacks.}

\fk{
In this subsection, we mainly exploit the robustness of GRFF against an iterative, gradient-based attack, called Iteratively, Least-Likely (Iter.L.L.) attack \cite{kurakin2016adversarial}, which is developed from a one-step, gradient-based attack, Fast Gradient Sign Method (FGSM, \cite{goodfellow2014explaining}).
By taking the gradient on input images and moving the images a small step along the direction of the sign of the gradient, FGSM creates the corresponding adversarial examples.
Based on this one-step FGSM, to acquire better attack performance, \citet{kurakin2016adversarial} proposed the iterative version, i.e., Iter.L.L. attack.
The steps for creating the adversarial example $\mathbf{x_{adv}}$ of an original example $\mathbf{x}$ can be summarized as follows,
\begin{equation}
\begin{split}
\mathbf{x_{adv}}^{(i+1)}&=
\mathrm{Clip}_{\mathbf{x},\epsilon}\{ \mathbf{x_{adv}}^{(i)} \},\\
\mathbf{x_{adv}}^{(i)}&\gets\mathbf{x_{adv}}^{(i)} - \alpha\cdot\mathrm{sign}(\nabla_\mathbf{x}L(\mathrm{model}(\mathbf{x_{adv}}^{(i)}),y_{ll})),\\
\mathbf{x_{adv}}^{(0)}&=\mathbf{x}.
\end{split}
\label{eq-ill}
\end{equation}
At the $i$-th iteration, Iter.L.L. attack computes the loss between the predicted label and the least-likely label $y_{ll}$,
moves the image a small step $\alpha$ along the direction of the sign of the gradient of the input,
and clips the pixel value of the perturbed image to a specified range $[\mathbf{x}_{i,j}-\epsilon,\mathbf{x}_{i,j}+\epsilon]$.
The step size is set as $\alpha=1$ by default.
The number of iterations is set as $\min\{\epsilon+4,1.25\epsilon\}$.}

\fk{
Iter.L.L. attack aims at linearizing the loss function around given fixed model parameters.
While the weights in GRFF are generated independently and thus are non-deterministic.
Hence, we evaluate the ability of GRFF in defending Iter.L.L. attack on MNIST \cite{lecun1998gradient} and have the following experiment design.
\begin{enumerate}
	\item Given an image $\mathbf{x}$ and noises $\mathbf{N}_{(1)}$ sampled from $\mathbb{P}_0$, GRFF can predict its label $\hat{y}_{(0)}$ and compute the loss between $\hat{y}_{(0)}$ and the least-likely label.
	\item By back propagation, the gradients on $\mathbf{x}$ are determined. Then the corresponding adversarial example $\mathbf{x_{adv}}$ is created according to Eq.\eqref{eq-ill}.
	\item GRFF can again take the original noises $\mathbf{N}_{(1)}$ as input and output a prediction $\hat{y}_{(1)}$ of $\mathbf{x_{adv}}$.
	\item By resampling independent noises $\mathbf{N}_{(2)}$ from $\mathbb{P}_0$, now GRFF can take the new noises $\mathbf{N}_{(2)}$ as input and output another prediction $\hat{y}_{(2)}$ on $\mathbf{x_{adv}}$.
\end{enumerate}
Based on the predicted results $\hat{y}_{(0)}$, $\hat{y}_{(1)}$ and $\hat{y}_{(2)}$ above, we could determine the corresponding unattacked accuracy $acc_{(0)}$, attacked accuracy with original noises $acc_{(1)}$, and attacked accuracy with resampled noises $acc_{(2)}$.
It is expected that the attacked accuracies $acc_{(1)}$ and $acc_{(2)}$ should be smaller than $acc_{(0)}$, and that $acc_{(2)}$ should be larger than $acc_{(1)}$ since resampling should be helpful in defending the Iter.L.L. attack.}

\begin{figure}[t]
	\centering
	\includegraphics[scale=0.5]{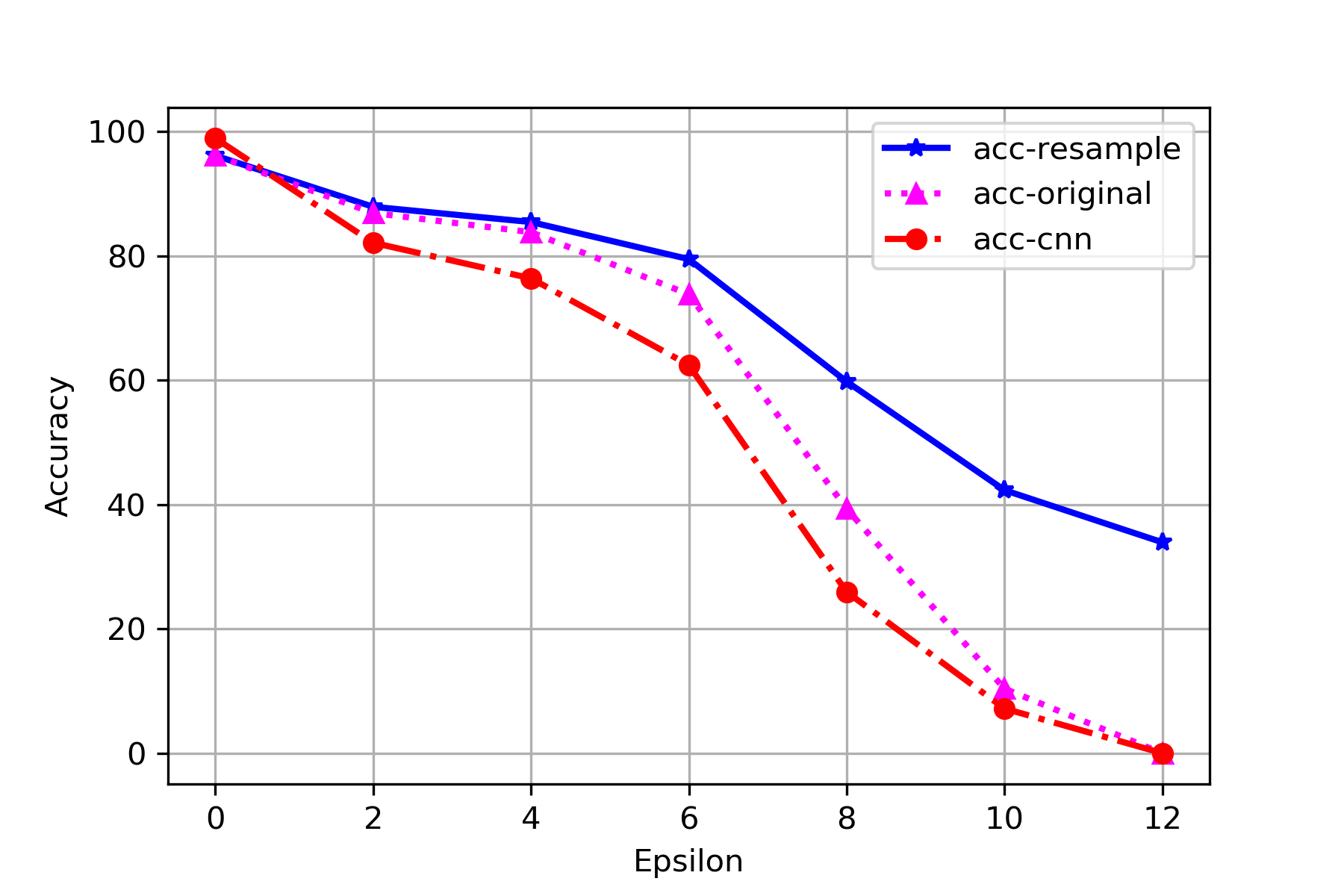}%
	\caption{Results of Iter.L.L. attack on GRFF and CNN. The blue solid curve and the magenta dotted curve denote the variations of $acc_{(2)}$ and $acc_{(1)}$ respectively. While $acc_{(0)}$ refers to the accuracy at $\epsilon=0$. The red dash-dot curve denotes the variation of the attacked accuracy of the CNN with cosine and sine activation functions.}
	\label{fig_attack_ill}
\end{figure}

\fk{
In this experiment, for the variant of GRFF, we also adopt a two-layer structure, the generators of which generate $16$ and $8$ weights of $5\times5$ size in the 1st and 2nd layers respectively.
The max-pooling operator takes a perception field of $2\times2$ size. 
The experiment results are shown in Fig.\ref{fig_attack_ill}.
By omitting the generators, we acquire a CNN with cosine and sine activation functions.
We also record the results of the Iter.L.L. attack on such a CNN in Fig.\ref{fig_attack_ill}.}

\begin{figure}[!t]
	\centering
	\includegraphics[scale=0.42]{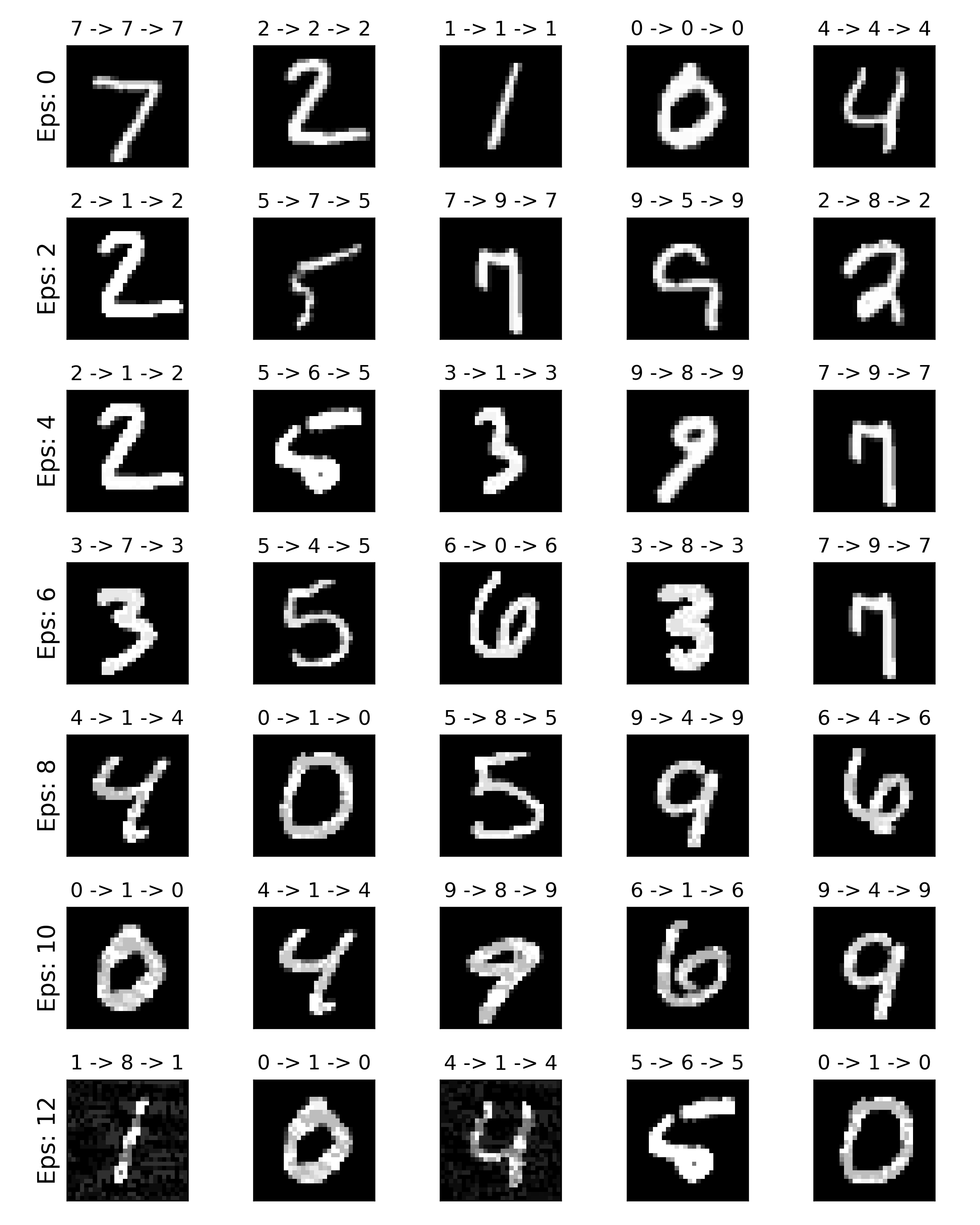}%
	\caption{An illustration of some adversarial examples of Iter.L.L attack. Different $\epsilon$ values correspond to the different rows. Among the 3 numbers at the top of each adversarial example, the first number denotes the true label, and the middle one denotes the prediction $\hat{y}_{(1)}$, and the last one denotes the prediction $\hat{y}_{(2)}$. These adversarial examples (except the first row) successfully fool GRFF without resampling. But GRFF successfully identifies these adversarial examples by resampling.}
	\label{fig_attack_adv_ex}
\end{figure}

\fk{
One can find that under Iter.L.L. attack, by resampling non-deterministic weights, $acc_{(2)}$ are generally larger than $acc_{(1)}$, indicating stronger adversarial robustness brought by the resampling mechanism of GRFF.
When $\epsilon$ equals to $12$ with $15$ iterations, all the adversarial examples successfully fool CNN and GRFF with original noises and result in zero accuracy.
While by resampling independent weights, GRFF successfully identifies nearly $40\%$ of these adversarial examples, which is a promising improvement.
Besides, both the attacked accuracies with resampled noises and original noises are higher than the attacked accuracy of CNN.
Therefore, due to the randomized resampling mechanism associated with GRFF, our model is able to alleviate the performance decrease brought by the Iter.L.L. attack.
It also shows superiority over the common CNN of the same structure in defending the Iter.L.L. attack.
Part of the adversarial examples are shown in Fig.\ref{fig_attack_adv_ex}.}

\section{Discussion and Conclusion}
\label{sec-conclusion}

In conclusion, 
we proposed a one-stage kernel learning approach,
which models some latent distribution of the kernel via a generative network based on the random Fourier features.
Not like the existing methods that learn the distribution by the target alignment and then train a linear classifier, we directly solved the ERM problem to jointly learn the features and the classifier for better generalization performance.
Further, such an end-to-end manner enables the model itself to extend to deeper layers, which leads to a multi-layer structure and implies a good kernel on features.
Besides, a progressively training strategy was proposed to efficiently train the multiple generators in an inverse, layer-by-layer order.
Empirical results illustrate the superiority of ML-GRFF in classification tasks over the classical two-stage, RFFs-based methods.
Meanwhile, GRFF enables us to resample independent and non-deterministic parameters from the generators.
Such parameter randomness is helpful in defending adversarial attacks and is empirically verified by the enhanced robustness against the adversarial examples.

One of the limitations of the proposed method is that its performance on large-scale image data and deeper networks is still restricted.
In our method, for image data, the generators try to learn the distributions on $3\times3$ or $5\times5$ weights.
However, in deeper network, like ResNet \cite{he2016deep}, a usual size of kernels in the convolution layer is $512\times512\times3\times3$.
The generator cannot include  $512\times512\times3\times3$ neurons in its output layer since it is a heavy burden for GPUs to compute the gradients.
Besides, more layers indicate more generators.
Training tens of, or even hundreds of generators still remains a difficult problem.

Since the parameter distribution is modeled by the generative networks, which cost huge memory and computation resources, future work will focus on how to design cheap and efficient mechanism to model the parameter randomness and to achieve better robustness.
Besides, based on the existing researches on the convergence of GANs \cite{liu2017approximation,farnia2020gans}, a theoretical guarantee on the convergence of the GRFF will be developed in the future.

\bibliographystyle{unsrtnat}  
\bibliography{reference}

% \begin{comment}
\clearpage

\begin{appendices}

\section{More discussions on ML-GRFF}
\label{sec:appendix}

In this section, discussions on ML-GRFF including over 2 layers are further raised.
Specifically, experiments on synthetic data are designed and executed to illustrate the feasibility of the training algorithm Alg.\ref{alg-train} even in the case of more than 2 layers.
We show that a two-layer structure is an empirically-optimal choice,  simultaneously considering the improved generalization performance and the difficulty of tuning hyper-parameters.
The designed experiments and results are outlined below.

\noindent\textbf{Settings}:

The experiments are executed on synthetic data, where $(\mathbf{x},y)$ follows $\mathbf{x}\sim\mathcal{N}(0,I_d)$ of dimension $d=80$ and $d=100$ with $y=\mathrm{sign}(||\mathbf{x}||^2-\sqrt{d})$.
The training set includes $10^4$ samples and the test set includes $10^3$ samples.
A validation set containing 20\% of the training data is partitioned to evaluate the model performance during training.

\begin{figure}[h]
	\centering
	\includegraphics[scale=0.42]{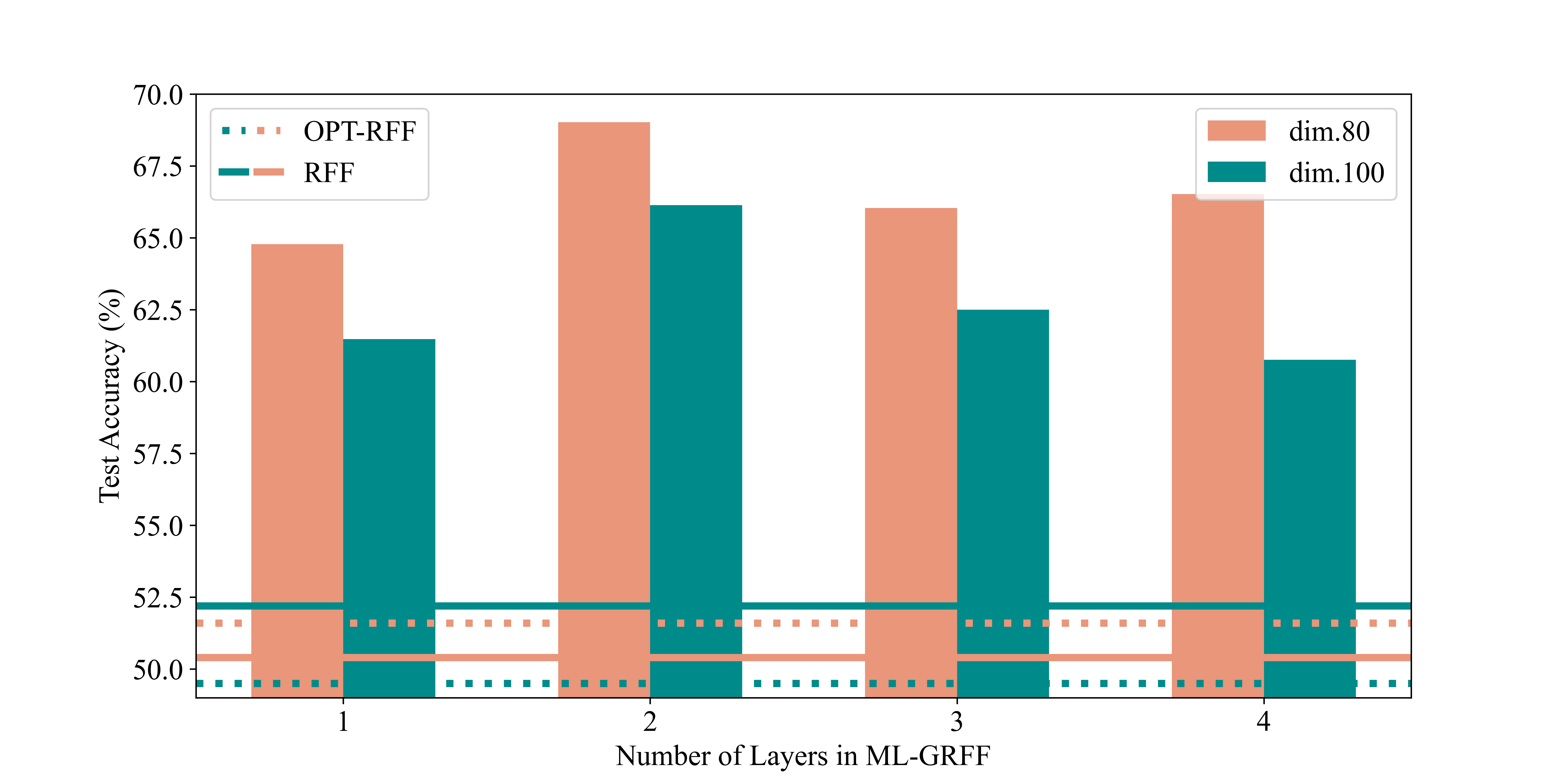}%
	\caption{Results on the test sets of synthetic data of ML-GRFF including $K$ layers, $K\in\big\{1,2,3,4\big\}$. The synthetic data is of 2 different dimensions, $d=80$ and $d=100$, denoted by distinct colors. The solid and dotted lines indicate the results of RFF and OPT-RFF respectively.}
	\label{fig:num-layers}
\end{figure}

For ML-GRFF with $K$ layers, we train multiple models w.r.t. $K\in\big\{1,2,3,4\big\}$.
Important hyper-parameters, e.g., numbers of sampled noises $\big\{D_1,D_2,\cdots,D_K\big\}$ and numbers of epochs $\big\{epo_1,epo_2,\cdots,epo_K\big\}$, are listed in Table \ref{tb-hyperparams}.
The experiments are repeated 5 times and the average results are recorded in Fig.\ref{fig:num-layers}.

\begin{table*}[h]
	\centering
	\caption{Hyper-parameters of training ML-GRFF with $K$ layers.}
	\label{tb-hyperparams}
% 	\resizebox{\textwidth}{!}{
	\begin{tabular}{c|c|c}
		\toprule[2pt]
		\# layers $K$  
		& $\big\{D_1,D_2,\cdots,D_K\big\}$ 
		& $\big\{epo_1,epo_2,\cdots,epo_K\big\}$	\\
		\midrule[0.5pt]
		$1$ & $\big\{256\big\}$ & $\big\{1000\big\}$ \\
		$2$ & $\big\{256,64\big\}$ 
		& $\big\{200,1000\big\}$ \\
		$3$ & $\big\{64,64,64\big\}$ 
		& $\big\{200,200,1000\big\}$ \\
		$4$ & $\big\{64,64,64,64\big\}$ 
		& $\big\{200,200,200,1000\big\}$ \\
		\bottomrule[2pt]
	\end{tabular}
% 	}
\end{table*}

\noindent\textbf{Results and discussions}:

As indicated in Fig.\ref{fig:num-layers}, as the number of layers increases over 2, the progressively training strategy could still guarantee the convergence of ML-GRFF and produce better results than RFF, OPT-RFF and single-layer GRFF.
On the other hand, the numbers of hyper-parameters of ML-GRFF also increase along with that of layers.
It remains a difficult problem to efficiently tune these hyper-parameters to always achieve significant generalization performance improvements in the case of more and more layers.
Based on the results in Fig.\ref{fig:num-layers}, empirically, a two-layer structure is an appropriate choice, and thus we focus on ML-GRFF of two layers in all the experiments.

\end{appendices}

\end{document}